\documentclass{article} % For LaTeX2e
\usepackage{iclr2026_conference,times}

\usepackage{amsmath,amsfonts,bm}

\def\eqref#1{equation~\ref{#1}}
\def\1{\bm{1}}
\def\vh{{\bm{h}}}

\def\vx{{\bm{x}}}
\def\vy{{\bm{y}}}
\def\vz{{\bm{z}}}

\def\mA{{\bm{A}}}

\def\mH{{\bm{H}}}
\def\mI{{\bm{I}}}

\def\mK{{\bm{K}}}

\def\mQ{{\bm{Q}}}

\def\mV{{\bm{V}}}
\def\mW{{\bm{W}}}

\DeclareMathAlphabet{\mathsfit}{\encodingdefault}{\sfdefault}{m}{sl}
\SetMathAlphabet{\mathsfit}{bold}{\encodingdefault}{\sfdefault}{bx}{n}

\usepackage{microtype}
\usepackage{graphicx}
\usepackage{subfigure}
\usepackage{booktabs} % for professional tables
\usepackage{makecell}

\usepackage{dashrule}
\usepackage{arydshln}

\usepackage[utf8]{inputenc} % allow utf-8 input
\usepackage[T1]{fontenc}    % use 8-bit T1 fonts

\usepackage{booktabs}       % professional-quality tables
\usepackage{amsfonts}       % blackboard math symbols
\usepackage{nicefrac}       % compact symbols for 1/2, etc.
\usepackage{microtype}      % microtypography
\usepackage{xcolor}         % colors

\usepackage{amsmath}
\usepackage{amssymb}
\usepackage{mathtools}
\usepackage{amsthm}
\usepackage{tabularray}
\usepackage{multirow} 
\usepackage{threeparttable}
\usepackage[most]{tcolorbox}
\usepackage{caption}
\usepackage{wrapfig}
\usepackage{amssymb}

\usepackage{hyperref}
\usepackage{url}

\usepackage{thm-restate}
\makeatletter
\def\thmheadbrackets#1#2#3{%
  \thmname{#1}\thmnumber{\@ifnotempty{#1}{ }\@upn{#2}}%
  \thmnote{ {\the\thm@notefont[#3]}}}
\makeatother

\newtheoremstyle{brakets}% Name
  {}% space above
  {}% space below
  {\itshape}% body font
  {}% indent
  {\bfseries}% head font
  {.}% punctuation after head
  { }% space after head (has to be space or dimension!)
  {\thmheadbrackets{#1}{#2}{#3}}% head spec

\newtheoremstyle{definitionbrakets}% Name
  {}                       % Space above
  {}                       % Space below
  {\normalfont}               % Body font (upright)
  {}                          % Indent amount
  {\bfseries}                 % Head font
  {.}                         % Punctuation after head
  { }                         % Space after head
  {\thmheadbrackets{#1}{#2}{#3}}% Head spec

\theoremstyle{brakets}

\theoremstyle{remark}

\theoremstyle{definitionbrakets}

\usepackage[textsize=tiny]{todonotes}

\newcount\Comments  % 0 suppresses notes to selves in text
\newcommand{\kibitz}[2]{\ifnum\Comments=1{\color{#1}{#2}}\fi}

\definecolor{tw}{rgb}{0.0, 0.0, 0.5}

\definecolor{usercolor}{RGB}{0, 0, 128}       % Navy Blue for User
\definecolor{cotcolor}{RGB}{128, 0, 0}        % Maroon for o1 CoT
\definecolor{outputcolor}{RGB}{0, 128, 0}     % Green for o1 Output
\newcommand{\shortn}{\textup{\texttt{-}}}
\newcommand{\shorte}{\textup{\texttt{=}}}

\newcommand{\rlm}{\textsc{Proxy LLM}}
\newcommand{\elm}{\textsc{Explanation LLM}}

\NewDocumentCommand{\idx}{o o o o}{\ensuremath{
{#1}
\IfValueT{#2}{\IfBlankTF{#2}{}{_{#2}}}
\IfValueT{#3}{\IfBlankTF{#3}{}{^{(#3)}}}
\IfValueT{#4}{\IfBlankTF{#4}{}{(#4)}}
}}

\title{Translate Policy to Language: Flow Matching Generated Rewards for LLM Explanations}

\author{%
 Xinyi Yang$^1 \thanks{The work was done while the author was a PhD student at Tsinghua.}$,
 Liang Zeng$^2 $  ,
 Heng Dong$^2$,
 Chao Yu$^3$,
 Xiaoran Wu$^4$, Huazhong Yang$^3$, \\
 \textbf{Yu Wang}$^3$,
 \textbf{Milind Tambe}$^5$,
 \textbf{Tonghan Wang}$^6 \thanks{Correspondence to tonghanwang1996@gmail.com}$,
   \vspace{.5em} 
  \\
  $^1$ AIPD, Tencent, Shenzhen, China
  \\
  $^2$ IIIS, Tsinghua University, Beijing, China 
  \\
  $^3$ EE, Tsinghua University, Beijing, China
  \\
  $^4$ CS, Tsinghua University, Beijing, China
  \\
  $^5$ SEAS, Harvard University, Cambridge, USA
  \\
  $^6$ College of AI, Tsinghua University, Beijing, China
}

\iclrfinalcopy % Uncomment for camera-ready version, but NOT for submission.

\begin{document}

\maketitle

\begin{abstract}
As humans increasingly share environments with diverse agents powered by RL, LLMs, and beyond, the ability to explain agent policies in natural language is vital for reliable coexistence. We introduce a general-purpose framework that trains explanation-generating LLMs via reinforcement learning from AI feedback, with distributional rewards generated by generative continuous normalizing flows (CNFs). CNFs capture the pluralistic and probabilistic nature of human judgments about explanations. Moreover, under mild assumptions, CNFs provably bound deviations from true human reward distributions when trained on noisy proxy rewards from LLMs. We design a specialized CNF architecture that selectively attends to linguistic cues in the decision context and explanations when generating rewards. Human and LLM evaluators find that our method delivers explanations that enable more accurate predictions of true agent decisions, exhibit greater logical soundness and actionability, and impose lower cognitive load than explanations trained with proxy LLM rewards or state-of-the-art RLHF and RLAIF baselines.

% Experiments on both RL and LLM tasks demonstrate that our method generates rewards significantly better than 

%Looking into the future, humans will likely inhabit an ecosystem shared by diverse intelligent agents powered by reinforcement learning (RL) models, large language models (LLMs), etc. Explaining their decisions in natural language is important to foster trust and enhance reliability of this ecosystem. In this paper we build a model-agnostic explanation generator based on an LLM. The technical novelty is that the rewards for training this LLM are generated from noise by a generative flow matching model. This model has a specially designed architecture with some hidden layers merged with an LLM to incorporate the linguistic cues of explanations into the reward generation process. Experiments on both RL and LLM tasks demonstrate that our method can generate dense and reasonable rewards while saving expensive human feedback, effectively enable agents to infer others' decisions from explanations, and even improve the success rate of the decisions in the original LLM task.
\end{abstract}

% (e.g., in an RL task, a decision is an RL action, and the context could be an MDP state)

\section{Introduction}

Intelligent agents, ranging from reinforcement learning (RL) agents~\citep{kiran2021deep,zhao2021consciousness,liu2024learning,qiu2024instructing}, large language models (LLMs)~\citep{yao2022react,shinn2024reflexion,wang2023voyager,kang2020incorporating}, to robotic systems~\citep{ismail2018survey}, are becoming increasingly intertwined with daily lives~\citep{survey2024wang}. To foster transparency, safety, and alignment, a promising avenue is communicating the reasoning behind actions or decisions generated by agent policies in natural language~\citep{lazaridou2016multi}. A good explanation provides a rationale that helps recover the underlying policy, enabling third parties to infer the decision.
Such explanations allow humans to comprehend decisions, provide meaningful feedback, and intervene when necessary, ultimately enhancing trust and reliability within intelligent ecosystems~\citep{cambria2023survey}.\looseness=-1

This effort is situated within the broader field of explainable AI and RL~\citep{arrieta2020explainable,carvalho2019machine,ehsan2019automated,gunning2017explainable,ras2018explanation,gilpin2018explaining}, 
%established areas encompassing
which encompasses methods such as causal inference and contrastive learning. Nevertheless, existing natural language explanation methods are often tailored to specific application scenarios, including self-driving~\citep{cai2024driving}, recommender systems~\citep{lubos2024llm}, stock prediction~\citep{koa2024learning}, robotics~\citep{lu2023closer}, and autonomous navigation~\citep{trigg2024natural}. To circumvent the extensive task- and agent-specific engineering, there is a need for a generalizable and task-agnostic explanation method. We achieve this by concealing agent decisions from an \elm, prompting it solely with decision contexts, and training it to recover a specific underlying policy and generate explanations that reflect that policy’s behavior. This approach forces the LLM to reason only from contextual information, independent of task or agent specifics, and thereby achieves generalizability.\looseness=-1

In reinforcement learning from human feedback (RLHF), such third-party evaluations are provided by humans. In explanation tasks, human feedback is inherently pluralistic~\citep{feng2024modular,sorensen2024value,kirk2024benefits,sorensen2024position} and probabilistic~\citep{lou2024uncertainty,dorka2024quantile}, as explanations are open-ended and subject to wide variation in individual judgments. This pluralism necessitates the costly collection of diverse human feedback in RLHF. Reinforcement learning from AI feedback (RLAIF)~\citep{bai2022constitutional,pace2024west,liu2024skywork} economizes human feedback, typically by adapting auxiliary LLMs to generate \emph{proxy rewards} that approximate human feedback. Although early work~\citep{li2024aligning} shows promise, current RLAIF approaches have not yet been rigorously studied on how to generate distributional rewards that explicitly manage proxy errors relative to human reward distributions.\looseness=-1

%We address these gaps by extending RLAIF with generative continuous normalizing flows (CNFs) \citep{chen2018neural,lipman2022flow} as proxy reward models, investigating how their ability in generating complex distributions from initial noise distributions can be exploited in delivering reward distributions for LLMs explanation training. 
% trained to approximate LLM-generated rewards
We close this gap by extending RLAIF with generative continuous normalizing flows (CNFs) \citep{chen2018neural,lipman2022flow} as reward models. CNFs map base noise distributions to complex distributions, a process we exploit to represent reward distributions for training the \elm. 
We train a CNF to fit reward samples generated by separate \rlm s. This method enjoys the theoretical benefit that, assuming the noise in these rewards generated by \rlm s has the same functional form as the CNF's base distribution (e.g., Gaussian), the CNF provably bounds its deviation from the underlying human reward distribution. Intuitively, the \rlm\ acts as a forward process that injects noise into human rewards, and the CNF learns its inverse, partially denoising and recovering the underlying human rewards. These analyses extend to all generative models based on ordinary differential equations (ODEs), including score-based diffusion models~\citep{song2021score}. Among them, we choose rectified flow (RF)~\citep{liu2022flow,albergo2023building} for its fast inference and high distributional fidelity. Appx.~\ref{appx:theorem} demonstrates the proof ideas of this theoretical benefit.

To operationalize these theoretical benefits, we condition the RF reward model on decision contexts and explanations. We learn to transform inputs of the RF model into \emph{flow tokens} that cross-attend to latent embeddings of decision contexts and explanations. The resulting attention summary informs the flow's output, allowing it to selectively integrate linguistic and contextual cues when assigning reward distributions.

We evaluate our method on both RL tasks ($\mathtt{SMAC}$,~\citep{samvelyan2019starcraft}) and LLM tasks ($\mathtt{MMLU}$ \citep{hendrycks2020measuring}, $\mathtt{MathQA}$ \citep{amini2019mathqa}). Removing the rectified flow and directly using rewards from proxy LLMs decreases performance by 6.9\%-12.4\%, demonstrating the effectiveness of the generated rewards. Moreover, our method outperforms $\mathtt{SFT}$, $\mathtt{RLHF}$~\citep{bai2022training}, $\mathtt{RLAIF}$~\citep{bai2022constitutional}, and reasoning baselines~\citep{trung2024reft} by 1.9\%-18.2\% across all tasks and backbone models. Human evaluations further confirm the high quality of our generated explanations, which receive the highest scores of accuracy, logicality, actionability, and cognitive load. \looseness=-1

\subsection{Related Works}

\textbf{LLM explanations}. 
% Previous work leveraging LLMs to generate explanations can be categorized into two approaches. For post-hoc natural language explanations, 
Methods such as AMPLIFY\citep{krishna2024post}, Self-Explain\citep{rajagopal2021selfexplain}, and Summarize and Score (SASC)~\citep{singh2023explaining} generate concise rationales based on agent decisions, sometimes accompanied by an explanation score to assess reliability. Unlike these methods, our approach trains LLMs to generate explanations without knowing agent decisions a priori. Chain-of-Thought (CoT) prompting~\citep{wei2022chain} is a widely adopted in-context learning technique that relies on step-by-step explanations or reasoning to enhance decision-making. Self-Taught Reasoner (STaR)~\citep{zelikman2022star} introduces an iterative refinement method, where a model improves its own explanations through self-generated rationales. While these methods are prompting-based and do not require additional training, optimization-based CoT methods like $\mathtt{CoT}$+$\mathtt{SFT}$~\citep{wei2022chain} and $\mathtt{ReFT}$~\citep{trung2024reft} have been developed. We compare against these optimization-based methods in our experiments for their proven powerful reasoning capability.

% optimization-based CoT methods like ReFT~\citep{trung2024reft} have  been developed. We compare against CoT and ReFT in our experiments as representative prompting-based and optimization-based baselines, respectively.
% demonstrating the potential of self-improvement for enhancing explanation quality
% Without knowing agent decisions, we train an LLM to generate informative and reliable explanations using a generative flow matching model.

% which seeks to provide insights into the causes of model decisions and make them understandable for humans
% While much of the research in this field has focused on supervised learning,

\textbf{RLHF}. Reinforcement learning from human feedback~($\mathtt{RLHF}$)~\citep{bai2022training, wang2023helpsteer, ouyang2022training, dong2024rlhf} is critical to aligning LLM behavior with human preferences such as helpfulness, harmlessness, and honesty~\citep{ganguli2022red, achiam2023gpt, team2023gemini}. An RL-based method uses a reward model~\citep{liu2024skywork} to approximate human preferences. Given a dataset of preferred versus rejected responses, a reward model can be trained using the standard Bradley-Terry model~\citep{bradley1952rank} with a pairwise ranking loss. The policy model (LLM) can then be fine-tuned by learned rewards via $\mathtt{PPO}$~\citep{schulman2017proximal}.
However, training a reward model can be costly. Direct preference learning ($\mathtt{DPO}$) \citep{rafailov2024direct} enables direct training on preference data, which can be adapted to different human utility models ($\mathtt{KTO}$,~\citep{ethayarajh2024kto}). We compare against $\mathtt{PPO}$, $\mathtt{DPO}$, and $\mathtt{KTO}$ in our experiments.

\textbf{RLAIF}. \citep{bai2022constitutional} introduces RLAIF, integrating LLM-generated preferences with human-defined principles for improved reward modeling. Subsequent studies, such as Skywork~\citep{liu2024skywork} and West-of-N~\citep{pace2024west}, further enhance this area by refining methods for curating preference data. While these approaches still partially rely on human annotations, other methods entirely exclude human data~\citep{wang2024self,wu2024meta}. In our experiments, we include comparisons with Skywork~\citep{liu2024skywork}, the leading RLAIF algorithm on the RewardBench leaderboard~\citep{lambert2024rewardbench}.
% with recent evidence suggesting synthetic datasets can match the performance of human annotations~\citep{}. 
% Our approach extends these synthetic data-driven methods by incorporating flow matching, which excels at modeling complex distributions. Comparative experiments demonstrate our method's strengths in tasks like interpreting RL decisions.

Several recent RLHF and RLAIF studies train \emph{distributional reward models} for LLM post-training. QRM~\citep{dorka2024quantile} employs quantile regression to estimate multimodal preference distributions. DPRM~\citep{li2024aligning} represents multiple preferences by a categorical distribution and approximates the target distribution via an optimal transport objective. In our setting, both methods would need to discretize explanation feedback--the probability that a third party can infer the true agent decision from explanations--and then fit a distribution on that discretized space. By contrast, our method seeks to model continuous feedback, learning distributions without discretization. URM~\citep{lou2024uncertainty} parameterizes Gaussian distributions to model disentangled human preference attributes, whereas our approach accommodates arbitrary reward distributions.

% While these approaches establish distributional reward modeling for human preference feedback, our work pioneers the extension of RLAIF through generative continuous normalizing flows as differentiable proxy reward models, enabling end-to-end preference density estimation beyond discrete human annotations.

% \textbf{Distributional reward modeling} has emerged as a critical advancement in RLHF/RLAIF paradigms, transitioning from deterministic scalar value estimation to probabilistic representations that explicitly characterize the spectrum of human preferences. 
% thereby better capturing the diversity of human values and addressing label noise. 

Please see Appx.~\ref{appx:related_work} for more related works on \textbf{Diffusion in Transformer}, \textbf{Cross-attention}, \textbf{Explainable AI/RL}, and \textbf{LLM Training in Explanations}.

\section{Preliminary}

% \label{sec:rw_rf}

\textbf{Rectified Flow}~\citep{liu2022flow,albergo2023building} emerges as a powerful generative model and has recently served as the basis for popular commercial tools like Stable Diffusion 3~\citep{stabilityAI2023}. Based on continuous normalizing flow (CNF)~\citep{chen2018neural,lipman2022flow}, it models a generative process as an Ordinary Differential Equation (ODE). Formally, a CNF transports an input $\vz_0\in \mathbb{R}^d$ to $\vz_t=\phi(t, \vz_0)$ at time $t\in[0,1]$ via the ODE:
\begin{align}
    \frac{d}{dt}\phi(t, \vz_0) = \varphi\left(t, \phi(t, \vz_0)\right).
\end{align}
$\phi:[0,1]\times \mathbb{R}^d\rightarrow\mathbb{R}^d$ is the \emph{flow}, and the \emph{vector field} $\varphi: [0,1]\times \mathbb{R}^d\rightarrow \mathbb{R}^d$ specifies the change rate of the state $\vz_t$. \citep{chen2018neural} suggests representing the vector field $\varphi$ with a neural network. The flow $\phi$ transforms an initial random variable $Z_0\sim p_0(\vz_0)$ to $Z_1\sim p_1(\vz_1)$ at time $t=1$. Rectified flow drives $\phi$ to follow the linear path in the direction $(Z_1-Z_0)$ as much as possible by optimizing the objective: $\min_\varphi \int_0^1 \mathbb{E}\left[\|(Z_1-Z_0)-\varphi(t, Z_t)\|^2\right]dt$, where $Z_t=t\cdot Z_1 + (1-t)\cdot Z_0$ is the linear interpolation of $Z_0$ and $Z_1$. Typically, the vector field network $\varphi$ is implemented as a U-Net~\citep{ronneberger2015unet} for image inputs or an MLP for vector inputs~\citep{wang2024diffusion}.

\textbf{Transformer}. The Transformer architecture~\citep{vaswani2017attention} is foundational to recent progress in LLMs~\citep{liu2024deepseek, zeng2024skywork,yang2024qwen2,team2023gemini}. For an input token sequence $x = (\idx[\vx][][1], \dots, \idx[\vx][][N])$, let \( \idx[E][][n][] = [e(\idx[\vx][][1]), \dots, e(\idx[\vx][][n])] \) denote the sequence of token embeddings up to position $n$, where $e(\cdot)$ is the token embedding function. A standard LLM generates its output by
\begin{align}
\idx[\mH][][n]=\textsc{Transformer}\left(\idx[E][][n]\right),\quad
M\left(\idx[\vy][][n+1] \mid \idx[\vx][][\leq n]\right)=\mW \idx[\vh][][n],\label{equ:token_logits}
\end{align}
where $\idx[\mH][][n] \in \mathbb{R}^{n \times d}$ denotes the last layer hidden states for the first $n$ tokens, with $d$ representing the hidden dimensionality. The last layer hidden state at position $n$ is $\idx[\vh][][n] = \idx[\mH][][n][] [n, :]$. We denote the output projection matrix by $\mW$, the generation logits by $M$, and the output tokens by $\vy$.

\section{Method}

Our goal is to train an \elm~$\pi_e(\theta_e)$ to recover a specific underlying policy and generate natural language explanations that reflect that policy’s decision of RL agents and LLMs given only context. In RL, the decision is an RL action, and the context is the trajectory of preceding states, actions, and rewards. In LLM tasks, the decision could be, for example, the selected option for a multiple-choice question, and the context is the question itself. Given a set $\mathcal{D}_e=\{(a_j,c_j)\}_{j=1}^{J}$ of decisions $a_j$ and their context $c_j$, we construct a prompt $x_j^e$ for $\pi_e(\theta_e)$: $\mathtt{Given}$ $\mathtt{\left[Context\ c_j\right].}$ $\mathtt{Please}$ $\mathtt{analyze}$ $\mathtt{the}$ $\mathtt{reasoning}$ $\mathtt{behind}$ $\mathtt{the}$ $\mathtt{agent}$ $\mathtt{decision}$ $\mathtt{based}$ $\mathtt{on}$ $\mathtt{the}$ $\mathtt{context}$. The \elm~then generates the corresponding explanation $y_j^e=\pi_e(x_j^e;\theta_e)$.

\begin{figure*}[t]
    \centering
    % \vspace{-0.5em}
    \includegraphics[width=\linewidth]{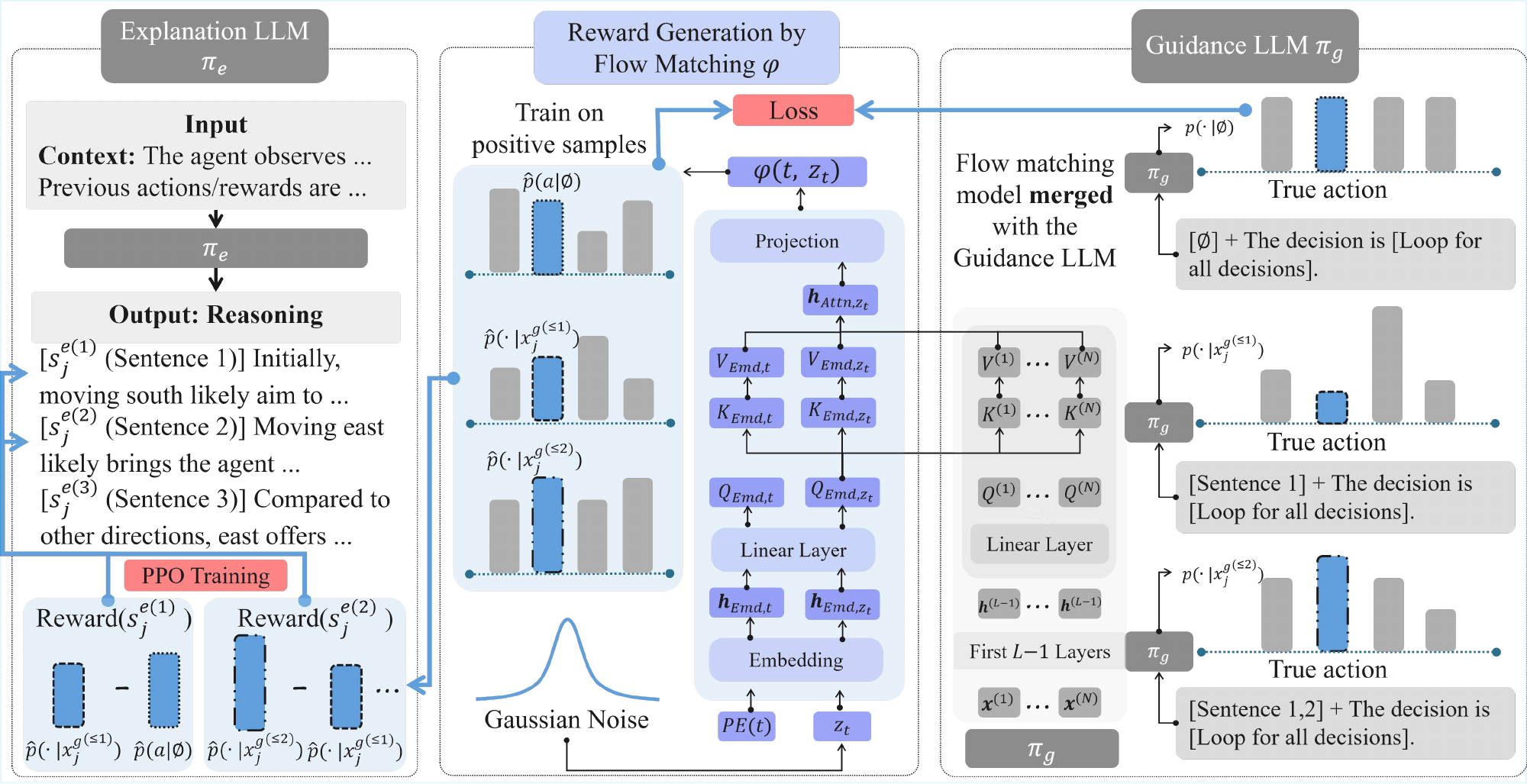}
    % \vspace{-1.5em}
    \caption{An overview of our method. (Left) We prompt an \elm~to generate reasoning about an agent decision based on the context information. Our focus is on whether a third party can infer the actual decision from this explanation. (Middle) We employ a rectified flow model $\varphi$ to generate rewards capturing how likely the actual decision appears as a plausible outcome after each sentence of the explanation. Per-sentence rewards for training the \elm~are the changes in the probability of the actual decision (highlighted in blue). (Right) The architecture and training of the rectified flow $\varphi$ are based on a \rlm. The \rlm~provides positive samples, where, with the context and explanation as input, it can produce a distribution $p$ that assigns the highest probability to the actual decision. The rectified flow $\varphi$ is trained to produce such distributions $p$, with a cross-attention layer in its middle that selectively leverages information from the \rlm~input, enabling generalization to negative samples.}
    \vspace{-0.5em}
    \label{fig:idea}
\end{figure*}

To prevent superficial explanations that simply restate the decision, we intentionally exclude the actual decision $a_j$ from the prompt $x_j^e$. For the generated content to be effective explanations, we train the \elm~using RL with rewards generated by a rectified flow $\varphi(\theta_\varphi)$. These rewards are probabilistic, reflecting the fact that human individuals have different opinions about the efficacy of the same explanation. Different explanation models correspond to different policies differentiated by training with distinct rewards. Fig.~\ref{fig:idea} illustrates our learning framework. The architecture and training of $\varphi(\theta_\varphi)$ are described in the following subsection.

% Fig.~\ref{fig:idea} illustrates our learning framework.

\subsection{Rectified flow reward model}

We aim at generating rewards that reflect how likely human individuals can infer correct decisions from explanations. To save costly human feedback~\citep{bai2022constitutional}, we use LLMs as proxy feedback providers when training the rectified flow reward model. Proxy LLM feedback may deviate from real human feedback. We empirically show the influence of these deviations as a baseline method $\mathtt{Proxy}$ $\mathtt{LLM}$ in Tab.~\ref{tab: ft_instruct} and formally analyze how the rectified flow manages these deviations in Sec.~\ref{sec:theory}.

Specifically, we query a \rlm~with the prompt $x_j^g$: $\mathtt{Given}$ $\mathtt{[Context\ c_j],}$ $\mathtt{the}$ $\mathtt{reasoning}$ $\mathtt{is\ [}y_j^e].$ $\mathtt{Thus,}$ $\mathtt{the}$ $\mathtt{decision\ is\ [a\ decision}$ $a\mathtt{]}$. For each candidate decision $a$, we measure the likelihood of it being a plausible outcome based on the token-level logits $M(a | x_j^g)$ (Eq.~\ref{equ:token_logits}), which indicate how likely the \rlm~is to output $a$ at the end of $x_j^g$. If $a$ spans multiple tokens, we compute the mean of their logits~\citep{yang2022logit}.

To capture the probabilistic nature of judgments on an explanation, we issue the same query for $K$ independent \rlm s, collecting logits $\{M_{j,k}(\cdot|x_j^g)\}_{k=1}^K$. The rectified flow $\varphi(\theta_\varphi)$ is trained by minimizing the following loss function:
\begin{align}
    \mathcal{L}_{\textsc{Flow}}(\theta_\varphi) &= \mathbb{E}_{\vz_0\sim Z_0, (c_j,y_j^e,M_{j,k})\sim \mathcal{D}_p, \vz_1=M_{j,k}, t\sim [0,1]} \left[\|(\vz_1-\vz_0)-\varphi(t,\vz_t|c_j,y_j^e; \theta_\varphi)\|^2\right].\label{equ:flow_loss}
\end{align}
where $\mathcal{D}_p=\{(c_j,y_j^e,\{M_{j,k}\}_{k=1}^K)\}_{j=1}^{J}$ is the training dataset and $\vz_t = t\cdot \vz_1 + (1-t)\cdot \vz_0$, for $t\in[0,1]$. The flow model transports each initial random sample $\vz_0\in\mathbb{R}^{|\mathcal{A}|}$ drawn from $Z_0$ to a logit vector of the same dimension, where $\mathcal{A}$ is the set of all candidate decisions. With initial samples from $Z_0$, $\varphi(\theta_\varphi)$ effectively constructs a distribution over logits. We define rewards as the logits corresponding to the ground-truth decision. Selecting them from the generated logits yields a corresponding distribution over rewards. We discuss how to use this reward distribution in Sec.~\ref{sec:train}.

%We treat different logits corresponding to the ground-truth decision as a reward distribution. By isolating these logits from the generated logit vectors, we thereby obtain a distribution over rewards.  

The choice of $Z_0$ is critical for the rectified flow to handle noise in the \rlm, as we will discuss in Sec.~\ref{sec:theory}.  

% After training, we use the output of the flow model $\varphi(\theta_\varphi)$ as rewards. 

\textbf{Rectified Flow Model Architecture}. As in a standard rectified flow model, the input to $\varphi$ includes (1) the current state $\vz_t\in\mathbb{R}^{|\mathcal{A}|}$, $t\in[0,1]$, with $\vz_0$ sampled from $Z_0$; and (2) a positional encoding $PE(t)$ corresponding to the ODE time $t\in[0,1]$. Apart from these, the inputs also include (3) context $c_j$ and explanations $y_j^e$. This requires that $\varphi$ can understand the linguistic cues in $c_j$ and $y_j^e$, which is beyond the capacity of typical rectified flow models based on fully-connected networks~\citep{wang2024diffusion} or U-Nets~\citep{liu2022flow}. To solve this problem, we propose to embed the rectified flow model $\varphi$ into an LLM.
% We demonstrate the specific network architecture in Figure~\ref{fig:idea} (Middle).
% $(\idx[\mQ][][N,L-1], \idx[\mK][][N,L-1], \idx[\mV][][N,L-1])$

\emph{Embedding}. The context $c_j$ and explanations $y_j^e$ are encoded by the~\elm, and we use its last layer latent states. ReLU-activated, layer-normalized MLPs $\varphi_{\textsc{Emb}}$ are used to project $\vz_t$ and $PE(t)$ to have the same dimension as the LLM tokens. $PE(t)$ and $\vz_t$ use two separate embedding MLPs: $\vh_{\textsc{Emb},t} = \varphi_{\textsc{Emb},t}(PE(t)),\vh_{\textsc{Emb},\vz_t} = \varphi_{\textsc{Emb},\vz_t}(\vz_t)$. The resulting embeddings, stacked as $\mH_{\textsc{Emb}}= (\vh_{\textsc{Emb},t},\vh_{\textsc{Emb},\vz_t})^\top\in\mathbb{R}^{2\times d}$, are called \emph{flow tokens}.

\emph{Cross-Attention}. We use the LLM's last layer weight matrices $(\mW_Q, \mW_K, \mW_V)$ to generate queries, keys, and values of the flow tokens: $(\mQ_{\textsc{Emb}}, \mK_{\textsc{Emb}}, \mV_{\textsc{Emb}}) = \mH_{\textsc{Emb}} (\mW_Q, \mW_K, \mW_V)$, which are concatenated with the last layer hidden states of the $c_j$ and $y_j^e$ to calculate cross-attention. More details are discussed in Appx.~\ref{appx:methods}.

\emph{Projector}. Define $\vh_{\textsc{Attn},\vz_t}$ as the latent state of the flow token $\vh_{\textsc{Emb},\vz_t}$ after cross-attention. This state has incorporated the ODE time $t$ and explanatory information through cross-attention. Progressing from this state, we use a fully-connected network $\varphi_{\textsc{Proj}}:\mathbb{R}^d\rightarrow\mathbb{R}^{|\mathcal{A}|}$ with ReLU activation and layer normalization to generate the vector field $\varphi= \varphi_{\textsc{Proj}}(\vh_{\textsc{Attn},\vz_t})$. We find that skip-layer connections are important for training stability, and we append the inputs $\vz_t$ and $t$ to the hidden layers of $\varphi_{\textsc{Proj}}$.

\subsection{Overall Training Procedure}\label{sec:train}

Once the rectified flow model $\varphi$ is trained, we use it to generate rewards for samples in $\mathcal{D}_e$. Concretely, we solve the ODE $d\vz_t = \varphi(t, \vz_t)dt$ starting from $\vz_0$ sampled from the initial distribution. An advantage of rectified flow is that the vector field $\varphi$ is encouraged to be straight lines, allowing efficient and accurate solution of the ODE in a few steps. The solution $\vz_1$ (at time $1$) is taken as the estimated logit $M_{\textsc{Flow}}(\cdot|x_j^g)=\vz_1=\vz_0+\int_0^1 \varphi(t,\vz_t)dt$.

Given different initial samples $\vz_0$, we might get different logits $M_{\textsc{Flow}}$, which reflect probabilistic opinions. We use the average logit of the true decision $M_{\textsc{Flow}}(a_j|x_j^g)$ as the reward. We then use PPO to train the \elm~to improve its explanations that achieve higher rewards from the rectified flow model. The rectified flow model and \elm~are trained alternately until convergence. The \elm~is fine-tuned using PPO with LoRA. The rectified flow model consists of a frozen LLM backbone and two trainable MLPs: $\varphi_{\textsc{Emb}}$ and $\varphi_{\textsc{Proj}}$.

\textbf{Implementation details}. In practice, we adopt rejection sampling~\citep{touvron2023llama} when training the flow model. Specifically, we only use samples where \rlm s assign the highest likelihood to the true decision: $a_j = \arg\max_a  M(a | x_j^g)$. To mitigate the influence of large-magnitude logit values, we apply a softmax activation to $ M(\cdot| x_j^g)$ before using it in the loss function (Eq.~\ref{equ:flow_loss}). Moreover, our model supports dense rewards. The logits change as we feed sentences in an explanation incrementally. We define per-sentence rewards as the changes of the true decision $a_j$'s logit after each newly added sentence. The sentence-level reward is used in addition to the sparse reward, where a reward of 1 is assigned only to the
last token of the explanation if the predicted decision is correct; otherwise, the reward is 0. More details are discussed in Appx.~\ref{appx:methods}. We study the effects of rejection sampling and sentence-level rewards in Sec.~\ref{sec:exp}. 

% For some decision-context pair $(a_j, c_j)$, the \rlm~is able to assign the highest probability to the actual decision $a_j$ compared to other alternatives by inferring from the explanation $\idx[s][e][1:k]$:

% , we use $\mathcal{D}_p$ to train a rectified flow model $\varphi$ to reproduce distribution $p(\cdot | \idx[x][g][k])$ from Gaussian noise.

% Given the training set $\mathcal{D}_e$, we first use the \elm~and \rlm s to construct the positive training set $\mathcal{D}_p$, which is used to train the rectified flow model $\varphi$ with learnable parameters in the embedding $\varphi_{\textsc{Emb}}$ and projection sub-network $\varphi_{\textsc{Proj}}$ by the loss in Eq.~\ref{equ:flow_loss}.

\section{Theoretical Analysis}\label{sec:theory}

In this section, we analyze the usage of a rectified flow model to approximate rewards.

Let $p(\cdot|y_j^e)$ be the unknown human reward distribution for the explanation $y_j^e$. As discussed, using LLMs as proxy human reward approximators might introduce errors. We first assume a Gaussian additive error model, and $p_{0}=\mathcal N(0,\sigma_{r}^{2}\mI_{|\mathcal{A}|})$ is the corruption noise. The LLM-generated distribution is given by the convolution $\hat{p}\;=\;p * p_{0}$. For rectified flow, we adopt an initial distribution of the same functional form as the corruption noise: $q_{0}=\mathcal N(0,\sigma^{2}\mI_{|\mathcal{A}|})$. Define the terminal distribution of the flow $\phi_{t}$ as $q_{1}=\phi_{1\sharp}q_{0}$. We can prove the following theorem.

\begin{restatable}{theorem}{flowerror}[Error bound for recovering the human reward distribution with a rectified flow] The flow is trained so that the fitting error to the proxy LLM reward distribution, measured in quadratic Wasserstein distance, satisfies $W_{2}(\hat{p},\,q_{1})\;\le\;\varepsilon$. $ p_{\textsc{flow}} := \phi_{1\sharp}p_{0}$ is the denoised candidate obtained by pushing $p_0$ through the learned flow. Errors relative to the true human distribution are bounded by
$$
W_{2}(p_{\textsc{flow}}, p)
\;\le\;
\varepsilon
\;+\;
L\,\sqrt{|\mathcal{A}|}\,\bigl|\sigma-\sigma_{r}\bigr|,
$$
where $L=\exp\bigl(\int_{0}^{1}\|\nabla \varphi_{t}\|_{\infty}\,dt\bigr)$ is the Lipschitz constant. In many architectures, $L\approx 1$.    
\end{restatable}
The proof in Appx.~\ref{appx:theorem} applies to any ODE-based generative models like score-based diffusion models. 
For other learning methods, such as previously proposed distributional reward modeling approaches~\citep{dorka2024quantile,lou2024uncertainty,li2024aligning}, we have the following lemma applicable when they learn from AI feedback.

\begin{restatable}{lemma}{error}[Error bound without a rectified flow] Without a rectified flow model, the error of a learning model approximating $\hat p$ relative to the true human distribution is bounded by
$$
W_{2}(\tilde{p}, p)
\;\le\;
\varepsilon'
\;+\;
\sqrt{|\mathcal{A}|}\,\bigl|\sigma_{r}\bigr|,
$$
where $\varepsilon'$ is the model's approximation error with respect to $\hat p$. Moreover, the inherent error in the proxy LLM reward distribution is $W_{2}(\hat p, p)=\sqrt{|\mathcal{A}|}\,\bigl|\sigma_{r}\bigr|$.
\end{restatable}

The major takeaway is that a rectified flow turns an unavoidable bias term $W_{2}(\hat p, p)=\sqrt{|\mathcal{A}|}\,\bigl|\sigma_{r}\bigr|$ into a controllable one as long as the initial distribution of the flow and the error model share the same functional form. A reasonable estimation of hyperparameters, e.g., $(\sigma\approx\sigma_{r})$, guarantees that a rectified model can well recover the true human reward distribution. We further provide a theoretical analysis in Appx.~\ref{appx:theorem} for the case where $q_0$ and $p_0$ follow different functional forms. Figure~\ref{fig:diffusion} in Appx.~\ref{appx:theorem} illustrates the proof ideas of our theorems.

% Moreover, when the training error satisfy $\varepsilon\le \sqrt d (\bigl|\sigma_{r}\bigr|-L\bigl|\sigma-\sigma_{r}\bigr|)$, then the flow model can even reduce inherent errors in \rlm s. This conclusion does not require the error model to be Gaussian. It holds as long as 

% Putting it all together: by explicitly modeling the corruption noise and leveraging the invertible-flow structure, a rectified flow turns an unavoidable bias term into a controllable one—and in the ideal hyperparameter setting removes it altogether—leaving only your network’s fitting error. This clean separation and the provable bound is what makes rectified flows strictly better than other deep‐learning approaches for recovering the true human reward distribution.

\section{Experiment}\label{sec:exp}

\textbf{Benchmark}. We evaluate our method on three testbeds: $\mathtt{(1)\ SMAC}$ (StarCraft Multiagent Challenge~\citep{samvelyan2019starcraft}) is an RL benchmark based on the real-time multi-agent strategy game StarCraft II that emphasizes micromanagement challenges. Explaining a decision here requires analyzing the status of ally and enemy units, such as their remaining health points and positions, and inferring their intentions from trajectories. 
$\mathtt{(2)\ AI2}$-$\mathtt{THOR}$ \citep{kolve2017ai2} is a widely used embodied AI benchmark in RL research \citep{li2024privacy, wu2021reinforcement}, featuring household activity scenarios that demand multi-step planning, multi-robot coordination, and interaction with dynamic environments. We use the \emph{Complex Tasks} split from SMART-LLM \citep{kannan2024smart} and expand it from 8 to 59 tasks. $\mathtt{(3)\ MMLU}$ (Massive Multitask Language Understanding \citep{hendrycks2020measuring}) is a multiple-choice QA benchmark for LLMs. We choose 4 challenging Professional Knowledge subsets (Professional Medicine (272 samples), Professional Law (1.53K samples), Professional Accounting (282 samples), and Professional Psychology (612 samples)). $\mathtt{(4)\ MathQA}$ \citep{amini2019mathqa} is a large-scale multiple-choice QA benchmark for math word problems, comprising 30K training samples and 3K testing samples. Please refer to Appx.~\ref{appx:experimental_setup} for more details.\looseness=-1
% 

% Follow the convention in LLM literature, We first supervised fine-tune the models on training datasets, and then train them use our method.
% To ensure a fair comparison, we reimplement these baselines and ablations using the same code base and datasets.

\textbf{Baselines}.
We evaluate our method against seven baseline families. \textbf{(1)\ Proxy LLM}. This baseline directly uses rewards generated by the \rlm~ to train the \elm. \textbf{(2)\ SFT} is a popular paradigm for adapting LLMs to downstream tasks. \textbf{(3)\ RLHF}. We consider $\mathtt{PPO}$~\citep{xu2024dpo}, $\mathtt{DPO}$~\citep{rafailov2024direct} and $\mathtt{KTO}$~\citep{ethayarajh2024kto}, using the TRL implementations~\citep{vonwerra2022trl}. \textbf{(4)\ Reasoning\ Frameworks}. We compare against $\mathtt{CoT}$+$\mathtt{SFT}$~\citep{wei2022chain} and $\mathtt{ReFT}$~\citep{trung2024reft}.
\textbf{(5) RLAIF}.  We report results for \textsc{Skywork} (Skywork-Reward-Llama-3.1-8B-v0.2)~\citep{liu2024skywork}, which ranks in the top two 8B models on the RewardBench leaderboard~\citep{lambert2024rewardbench}. \textbf{(6)\ DeepSeek}. We compare with DeepSeek-R1-Distill-Llama-8B~\citep{guo2025deepseek}. \textbf{(7)\ o3-mini}. We directly use $\mathtt{o3}$-$\mathtt{mini}$~\citep{openai_o3mini_system_card_2025} for evaluation without fine-tuning. \looseness=-1

%and provides the backbone and preference data for the remaining top-five models
% They require high-quality training data.
% a distilled version of DeepSeek-R1

%For example, on MMLU, KTO trained on our synthetic dataset achieves an accuracy of 62.5\%, outperforming the 58.6\% ACC obtained with an established instruction-tuning dataset~\citep{ethayarajh2024kto}.

% As the MathQA dataset is well-suited for step-by-step reasoning in mathematical problems, it is used for evaluating reasoning methods. In contrast, the SMAC and MMLU datasets are used for comparisons involving SFT and RLHF baselines.

\textbf{Ablations}. \textbf{(1)} $\mathtt{Ours\ w/o\ Attn}$. We remove the cross-attention layer in the reward model $\varphi$ and use a fully-connected network to generate the vector field. \textbf{(2)} $\mathtt{Ours}$ $\mathtt{(Sparse\ Reward)}$. We use a single reward for the last token of the explanation instead of sentence-level rewards.
\textbf{(3)} $\mathtt{PPO}$ $\mathtt{(Sentence\ Reward)}$. We apply sentence-level rewards to $\mathtt{PPO}$. \textbf{(4)} $\mathtt{Ours\ w/o\ Flow}$ is the same as the baseline $\mathtt{Proxy}$ $\mathtt{LLM}$.

% combine $\vh_{\textsc{Emb},t}$, $\vh_{\textsc{Emb},\vz_t}$ and the hidden state extracted from the  A 

\textbf{Models and Training}. By default, we use Llama-3.1-8B-Instruct~\citep{dubey2024llama} as the backbone of the \elm. We also test different backbone models, including Llama-3.1-8B, Qwen2.5-7B-Instruct~\citep{yang2024qwen2} and Gemma-2-2B-It~\citep{lieberum2024gemma} to assess the robustness of our method. 
% {\color{red}Additionally, to verify the effectiveness of the instruct-tuned model, we further test  as both the \elm~and the \rlm. }

%for generating step-by-step explanations
All methods except $\mathtt{o3}$-$\mathtt{mini}$ are trained on a shared synthetic dataset generated by o1-mini~\citep{jaech2024openai} and verified by human evaluators. We prompt o1-mini to generate paired \emph{Preferred explanations} given ground-truth decisions and \emph{Rejected explanations} given randomly selected incorrect decisions. The training dataset for $\mathtt{CoT}$+$\mathtt{SFT}$ and $\mathtt{ReFT}$ is further enhanced by asking o1-mini to adopt the $\mathtt{CoT}$ paradigm. We didn't directly use established instruction-tuning datasets because they provide preference data for answers, not explanations. 
To evaluate the quality of our synthetic
dataset, 
\begin{wraptable}[7]{r}{4.3cm}
\centering
\vspace{-1em}
\caption{ Human evaluation of our constructed datasets.}\label{tab: human_exp_data}
\vspace{-0.8em}
\begin{tabular}{crcr}
        \toprule
      \multicolumn{2}{c}{Preferred$\uparrow$}&
      \multicolumn{2}{c}{Rejected$\downarrow$}\\
         \cmidrule(lr){1-2}
        \cmidrule(lr){3-4}
        \multicolumn{2}{c}{3.53(0.53)} & 
        \multicolumn{2}{c}{2.26(0.76)}\\
        \toprule
    \end{tabular}
\end{wraptable}
12 participants (2 females, 10 males, aged 20–40) are invited to rate 5\% 
randomly chosen samples from each benchmark and their corresponding explanations on a 1–4 scale based on decision inferability. As shown in Tab.~\ref{tab: human_exp_data}, the results (score mean ± variance) indicate that preferred samples significantly enhance human inference accuracy compared to rejected samples, confirming the reliability of our synthetic dataset. 

\begin{figure}[t]
% \captionsetup{justification=centering}
 \centering
    \subfigure[\label{fig: positive}\scriptsize{Positive Samples}]
        {\centering
        {\includegraphics[height=2.3cm]{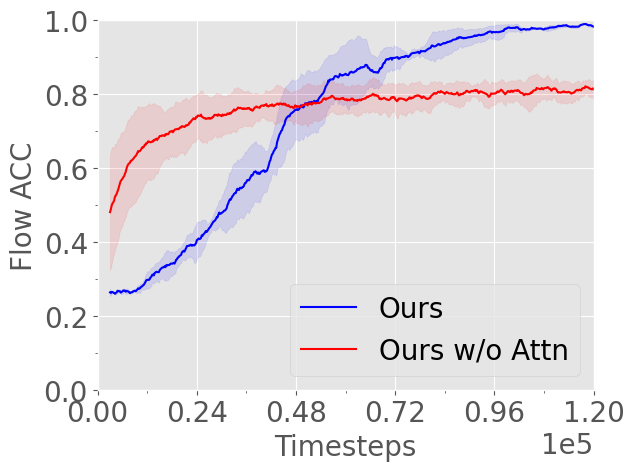}
            }
    }
    \subfigure[\label{fig: negative}\scriptsize{Negative Samples}]
        {\centering
        {\includegraphics[height=2.3cm]{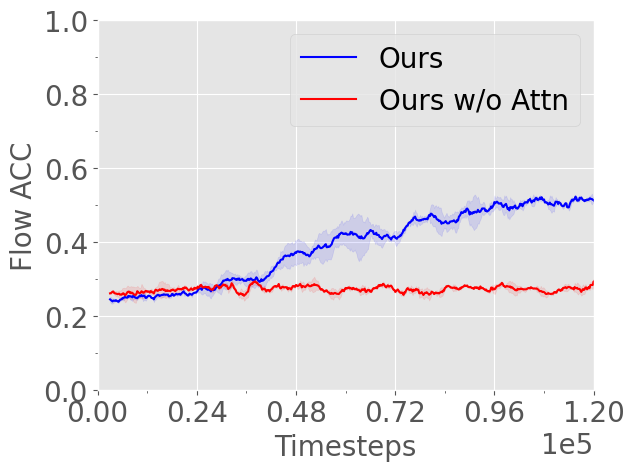}
    	}
    }
    \subfigure[\label{fig: true_false_negative}\scriptsize{True/False Negative Samples}]
        {\centering
        {\includegraphics[height=2.3cm]{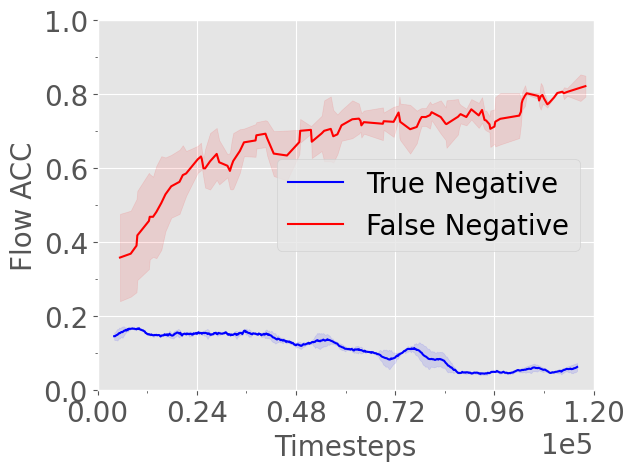}
    	}
    }
    \subfigure[\label{fig: elm_training}]
        {\centering
        {\includegraphics[height=2.3cm]{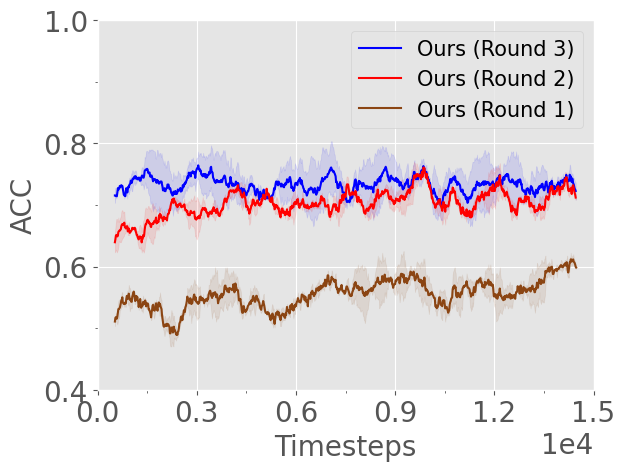}
    	}
    }

     \vspace{-2mm}
    \caption{(a-b) Accuracy of the rectified model $\varphi$ on unseen test samples. Flow ACC is reported as the percentage of samples for which the flow model  $\varphi$ assigns the highest logit (reward) to the ground-truth decision. \textbf{(c) Our rectified reward model learns to assign low rewards to true negatives---explanations that lead to wrong decisions}---and high rewards to false negatives. Rewards are positively related to accuracy, represented by the y-axis. (d) \elm~accuracy increases with each training round. Each training step corresponds to processing a single training sample.\looseness=-1} 
    \label{fig: rectified_flow_training}
    % \vspace{-2em}
\end{figure}

All methods except $\mathtt{o3}$-$\mathtt{mini}$ undergo three epochs of SFT on the preferred subset. Baselines ($\mathtt{PPO}$, $\mathtt{DPO}$, $\mathtt{KTO}$, $\mathtt{Skywork}$ and $\mathtt{ReFT}$) then perform preference optimization with the full preferred/rejected pairs. For our approach, the \elm~and the rectified flow model are trained iteratively for two rounds. As shown in Fig.~\ref{fig: elm_training}, additional rounds provide only marginal gains. For the rectified flow model, we choose $\sigma=1$. We use $K=3$ independent \rlm s: Llama-3.1-8B-Instruct, Qwen2.5-7B-Instruct and Gemma-2-2B-It. Complete training details and hyperparameter settings can be found in Appx.~\ref{appx:experimental_setup}.

%Our method, baselines and ablations are trained on a shared synthetic dataset. The training dataset is constructed by OpenAI o1-mini~\citep{jaech2024openai} and verified by human evaluators. We prompt OpenAI o1-mini to generate paired \emph{Preferred explanations} for ground-truth decisions and \emph{Rejected explanations} for randomly selected incorrect decisions. The training dataset for $\mathtt{CoT}$+$\mathtt{SFT}$ and $\mathtt{ReFT}$ is further enhanced by asking OpenAI o1-mini to adopt CoT paradigm. We didn't directly use established instruction-tuning datasets because they provide preference data for answers, not explanations.

% All methods undergo three epochs of SFT on the preferred subset. Baselines ($\mathtt{PPO}$, $\mathtt{DPO}$, $\mathtt{KTO}$, $\mathtt{Skywork}$ and $\mathtt{ReFT}$) then perform preference optimization with the full preferred/rejected pairs. 
% Consequently, these datasets can't perform very well in our tasks. 

% A negative sample from $\mathtt{MMLU}$. Although the explanation is correct, the \rlm~produces a wrong distribution. The rectified flow is not trained on sample, but can give a correct distribution. Specially, the explanation hinges on "the egg roll was present for a substantial time". Whereas the \rlm~missed this cue, the rectified flow model can generalize and predict correctly.

\textbf{Evaluation and Metrics}. 
For LLM evaluation, an explanation generated by the \elm~is considered effective if GPT-4o correctly infers the ground-truth decision based on it, with accuracy (ACC) measured as the proportion of effective explanations. We also consider the success rate of the whole task as the evaluation metric. For human evaluation, we recruit 45 participants (30 males, 15 females, aged 20-40) from diverse backgrounds (engineering, chemistry, machine learning, and economics). Each participant reviews 3 randomly selected samples from the benchmark and predicts the ground-truth labels. We report prediction accuracy and the percentage of participants rating each method the highest logical soundness, the highest actionability, and the lowest cognitive load. More details can be found in Appx.~\ref{appx:experimental_setup}.

% We consider the flow model produces an effective reward if the highest logit in the logit vector generated by the flow model corresponds to the ground-truth decision. The Flow ACC is calculated by the percentage of effective reward generation.

%We consider a reward generated by the flow model to be effective if the highest logit in the logit vector generated by the flow model corresponds to the ground-truth decision. Flow ACC is computed as the percentage of such effective rewards.

%and area under the receiver operating characteristic curve (AUC) as evaluation metrics.

\subsection{Comparisons against Baselines}

\textbf{LLM Evaluation}. Tab.~\ref{tab: ft_instruct} reports the results. For $\mathtt{SMAC}$, $\mathtt{MMLU}$, and $\mathtt{MathQA}$ benchmark, compared to $\mathtt{Proxy}$ $\mathtt{LLM}$, our method achieves 12.4\%, 6.9\%, and 11.0\% higher ACC on these datasets, respectively. These results support our theoretical analyses that rewards from \rlm s yield ineffective optimization, while the rectified flow model better handles noisy proxy rewards. Among $\mathtt{RLHF}$ baselines, $\mathtt{KTO}$ and $\mathtt{DPO}$ perform relatively well due to strong human utility models. In Reasoning Frameworks, although $\mathtt{ReFT}$ improves ACC to 72.2\%, 74.3\%, and 76.3\% on these datasets, it still lags behind our method. $\mathtt{Skywork}$, $\mathtt{DeepSeek}$ and $\mathtt{o3}$-$\mathtt{mini}$ also underperform our method. 
The performance of our method remains significantly better than baselines in the challenging $\mathtt{AI2}$-$\mathtt{THOR}$ benchmark. It improves the success rate by 7.4\%, 6.0\%, and 2.5\% over $\mathtt{KTO}$, $\mathtt{ReFT}$ and $\mathtt{o3}$-$\mathtt{mini}$, respectively.
In summary, our method consistently surpasses all baselines, demonstrating its effectiveness in generating more reasonable explanations that support inferring correct decisions.

\begin{table} [h]
    \caption{Accuracy results of our method and baselines on ${\mathtt{SMAC}}$, ${\mathtt{MMLU}}$ and ${\mathtt{MathQA}}$. Success rate of our method and baselines on ${\mathtt{AI2}}$-${\mathtt{THOR}}$.
    % ${\mathtt{Proxy}}$ ${\mathtt{LLM}}$ is the same as $\mathtt{Ours\ w/o\ Flow}$.
    }
    \vspace{-0.8em}
    \centering
    \scalebox{0.82}{
    \begin{tabular}{crcrcrcrcrcrcrcrcrcrcrcr}
        \toprule

        \multicolumn{2}{c}{\multirow{2}{*}{\makecell{${\mathtt{Datasets}}$}}}&
       \multicolumn{2}{c}{\multirow{2}{*}{\makecell{${\mathtt{Ours}}$}}}&
       \multicolumn{2}{c}{\multirow{2}{*}{\makecell{${\mathtt{SFT}}$}}}&
        \multicolumn{2}{c}{\multirow{2}{*}{\makecell{${\mathtt{Proxy}}$\\${\mathtt{LLM}}$}}}&
        \multicolumn{6}{c}{${\mathtt{RLHF}}$}&
        \multicolumn{4}{c}{$\mathtt{Reasoning}$}&
        \multicolumn{2}{c}{${\mathtt{RLAIF}}$}&
        \multicolumn{2}{c}{\multirow{2}{*}{\makecell{${\mathtt{DeepSeek}}$}}}&
        \multicolumn{2}{c}{\multirow{2}{*}{\makecell{${\mathtt{o3-}}$\\${\mathtt{mini}}$}}}\\

        \cmidrule(lr){9-14}
        \cmidrule(lr){15-18}
        \cmidrule(lr){19-20}
        
        \multicolumn{2}{l}{} &
        \multicolumn{2}{c}{}&
        \multicolumn{2}{c}{}&
        \multicolumn{2}{c}{}&
        \multicolumn{2}{c}{${\mathtt{PPO}}$}&
        \multicolumn{2}{c}{${\mathtt{DPO}}$}&
        \multicolumn{2}{c}{${\mathtt{KTO}}$}&
        \multicolumn{2}{c}{$\mathtt{CoT}$+$\mathtt{SFT}$}&
        \multicolumn{2}{c}{${\mathtt{ReFT}}$}&
        \multicolumn{2}{c}{${\mathtt{Skywork}}$}\\

         % \midrule

         \cmidrule(lr){1-2}
        \cmidrule(lr){3-4}
        \cmidrule(lr){5-6}
        \cmidrule(lr){7-8}
           \cmidrule(lr){9-10}
        \cmidrule(lr){11-12}
        \cmidrule(lr){13-14}
        \cmidrule(lr){15-16}
        \cmidrule(lr){17-18}
           \cmidrule(lr){19-20}
        \cmidrule(lr){21-22}
        \cmidrule(lr){23-24}

        % \cmidrule(lr){5-14}

        % \cmidrule(lr){1-2}
        % \cmidrule(lr){3-4}
        % \cmidrule(lr){5-6}
        % \cmidrule(lr){7-8}
        % \cmidrule(lr){9-10}
        % \cmidrule(lr){11-12}
        % \cmidrule(lr){13-14}

        % & \multicolumn{2}{l}{100\%} & \multicolumn{2}{l}{100\%} & \multicolumn{2}{l}{100\%} & \multicolumn{2}{l}{100\%} \\
        % \multicolumn{2}{c}{} & \multicolumn{2}{l}{}

        \multicolumn{2}{l}{${\mathtt{SMAC}}$} & 
        \multicolumn{2}{c}{\textbf{0.764}} &
        \multicolumn{2}{c}{0.582} & 
        \multicolumn{2}{c}{0.640} &
        \multicolumn{2}{c}{0.653} & 
        \multicolumn{2}{c}{0.713} & 
        \multicolumn{2}{c}{0.721} & 
        \multicolumn{2}{c}{0.642} & 
        \multicolumn{2}{c}{0.722} & 
        \multicolumn{2}{c}{0.692} & 
        \multicolumn{2}{c}{0.658}& 
        \multicolumn{2}{c}{0.455}  \\
        
       \cmidrule(lr){1-2}
        \cmidrule(lr){3-4}
        \cmidrule(lr){5-6}
        \cmidrule(lr){7-8}
           \cmidrule(lr){9-10}
        \cmidrule(lr){11-12}
        \cmidrule(lr){13-14}
        \cmidrule(lr){15-16}
        \cmidrule(lr){17-18}
           \cmidrule(lr){19-20}
        \cmidrule(lr){21-22}
        \cmidrule(lr){23-24}

        \multicolumn{2}{l}{${\mathtt{MMLU}}$
        } & 
        \multicolumn{2}{c}{\textbf{0.772}} &
        \multicolumn{2}{c}{0.689} & 
        \multicolumn{2}{c}{0.703} &
        \multicolumn{2}{c}{0.723}  & 
        \multicolumn{2}{c}{0.746} & 
        \multicolumn{2}{c}{0.753} & 
        \multicolumn{2}{c}{0.705} & 
        \multicolumn{2}{c}{0.743} & 
        \multicolumn{2}{c}{0.737} & 
        \multicolumn{2}{c}{0.721}& 
        \multicolumn{2}{c}{0.707}\\

        \cmidrule(lr){1-2}
        \cmidrule(lr){3-4}
        \cmidrule(lr){5-6}
        \cmidrule(lr){7-8}
           \cmidrule(lr){9-10}
        \cmidrule(lr){11-12}
        \cmidrule(lr){13-14}
        \cmidrule(lr){15-16}
        \cmidrule(lr){17-18}
           \cmidrule(lr){19-20}
        \cmidrule(lr){21-22}
        \cmidrule(lr){23-24}

        \multicolumn{2}{l}{${\mathtt{MathQA}}$} & 
        \multicolumn{2}{c}{\textbf{0.804}} &
        \multicolumn{2}{c}{0.667} & 
        \multicolumn{2}{c}{0.694} &
        \multicolumn{2}{c}{0.719} & 
        \multicolumn{2}{c}{0.753} & 
        \multicolumn{2}{c}{0.758} & 
        \multicolumn{2}{c}{0.712} & 
        \multicolumn{2}{c}{0.763} & 
        \multicolumn{2}{c}{0.729} & 
        \multicolumn{2}{c}{0.718}& 
        \multicolumn{2}{c}{0.739}  \\

        \midrule
        \midrule

        \multicolumn{2}{l}{${\mathtt{AI2}}$-${\mathtt{THOR}}$
        } & 
        \multicolumn{2}{c}{\textbf{0.702}} &
        \multicolumn{2}{c}{0.447} & 
        \multicolumn{2}{c}{$\backslash$} &
        \multicolumn{2}{c}{$\backslash$}  & 
        \multicolumn{2}{c}{$\backslash$} & 
        \multicolumn{2}{c}{0.628} & 
        \multicolumn{2}{c}{$\backslash$} & 
        \multicolumn{2}{c}{0.642} & 
        \multicolumn{2}{c}{0.483} & 
        \multicolumn{2}{c}{0.476}& 
        \multicolumn{2}{c}{0.677}\\

        \toprule
    \end{tabular}}
    % \begin{tablenotes}
    %     \item[1] xxx.
    % \end{tablenotes}
    \label{tab: ft_instruct}
    \vspace{-1em}
\end{table}

\textbf{Human Evaluation}. In LLM evaluation, $\mathtt{DPO}$ and $\mathtt{KTO}$ lead the $\mathtt{RLHF}$ baselines, $\mathtt{ReFT}$ tops the reasoning frameworks. Therefore, for human evaluation, we compare our method against $\mathtt{DPO}$ and $\mathtt{ReFT}$. Results on MathQA are shown in Tab.~\ref{tab: human_exp_results}. Our method achieves the highest accuracy, exceeding the strongest baseline by 25.7\%. Around 60\% of participants judge our explanations to be the most logically sound and the least cognitively demanding, indicating improved decision comprehension.

\begin{minipage}{\textwidth}

\begin{minipage}[ht!]{0.45\textwidth}
\makeatletter\def\@captype{table}
    \caption{Human evaluations on $\mathtt{MathQA}$, where Cognitive denotes cognitive load.}
        \vspace{-0.8em}
    
    \centering
      \scalebox{0.85}{
     \setlength\tabcolsep{1.5mm}{\begin{tabular}{crcrcrcrcr}
        \toprule
        
        \multicolumn{2}{l}{Methods} &
      \multicolumn{2}{c}{ACC}&
      \multicolumn{2}{c}{Logic}
      &\multicolumn{2}{c}{Actionable}
      &\multicolumn{2}{c}{Cognitive}\\

         \cmidrule(lr){1-2}
        \cmidrule(lr){3-4}
        \cmidrule(lr){5-6}
        \cmidrule(lr){7-8}
        \cmidrule(lr){9-10}

       \multicolumn{2}{l}{${\mathtt{Ours}}$}& 
        \multicolumn{2}{c}{\textbf{0.892}} & 
        \multicolumn{2}{c}{\textbf{0.60}}& 
        \multicolumn{2}{c}{\textbf{0.46}}& 
        \multicolumn{2}{c}{\textbf{0.60}}\\
         \midrule
        \multicolumn{2}{l}{${\mathtt{DPO}}$}& 
        \multicolumn{2}{c}{0.591} & 
        \multicolumn{2}{c}{0.17}& 
        \multicolumn{2}{c}{0.28}& 
        \multicolumn{2}{c}{0.18}\\
         \midrule
        \multicolumn{2}{l}{${\mathtt{ReFT}}$}& 
        \multicolumn{2}{c}{0.635} & 
        \multicolumn{2}{c}{0.23}& 
        \multicolumn{2}{c}{0.26}& 
        \multicolumn{2}{c}{0.22}\\
        % \midrule
        % \multicolumn{2}{l}{${\mathtt{Skywork}}$}& 
        % \multicolumn{2}{c}{0.619} & 
        % \multicolumn{2}{c}{--}& 
        % \multicolumn{2}{c}{--}& 
        % \multicolumn{2}{c}{--}\\

        \toprule
    \end{tabular}}}
\vspace{-0.8em}
\label{tab: human_exp_results}

\end{minipage}
\begin{minipage}[ht!]{0.55\textwidth}
\makeatletter\def\@captype{table}
\caption{Accuracy of our method and baselines with rejection sampling.}
    \vspace{-0.8em}
    \centering
    \scalebox{0.9}{

    \begin{tabular}{crcrcrcrcrcr}
        \toprule

        \multicolumn{2}{l}{Datasets} &
        \multicolumn{2}{c}{${\mathtt{Ours}}$}&
        \multicolumn{2}{c}{${\mathtt{PPO}}$}&
        \multicolumn{2}{c}{${\mathtt{DPO}}$}&
        \multicolumn{2}{c}{${\mathtt{KTO}}$}&
        \multicolumn{2}{c}{${\mathtt{Skywork}}$}\\

         % \midrule

         \cmidrule(lr){1-2}
        \cmidrule(lr){3-4}
        \cmidrule(lr){5-6}
        \cmidrule(lr){7-8}
           \cmidrule(lr){9-10}
        \cmidrule(lr){11-12}

        \multicolumn{2}{l}{${\mathtt{SMAC}}$} & 
        \multicolumn{2}{c}{\textbf{0.764}} & 
        \multicolumn{2}{c}{0.642} & 
        \multicolumn{2}{c}{0.704} & 
        \multicolumn{2}{c}{0.708} & 
        \multicolumn{2}{c}{0.713}  \\
        
       \cmidrule(lr){1-2}
        \cmidrule(lr){3-4}
        \cmidrule(lr){5-6}
        \cmidrule(lr){7-8}
           \cmidrule(lr){9-10}
        \cmidrule(lr){11-12}

        \multicolumn{2}{l}{${\mathtt{MMLU}}$} & 
        \multicolumn{2}{c}{\textbf{0.772}} & 
        \multicolumn{2}{c}{0.715}  & 
        \multicolumn{2}{c}{0.738} & 
        \multicolumn{2}{c}{0.759} & 
        \multicolumn{2}{c}{0.720}  \\

        \cmidrule(lr){1-2}
        \cmidrule(lr){3-4}
        \cmidrule(lr){5-6}
        \cmidrule(lr){7-8}
           \cmidrule(lr){9-10}
        \cmidrule(lr){11-12}

        \multicolumn{2}{l}{${\mathtt{MathQA}}$} & 
        \multicolumn{2}{c}{\textbf{0.804} }  & 
        \multicolumn{2}{c}{0.732}  & 
        \multicolumn{2}{c}{0.765} & 
        \multicolumn{2}{c}{0.773} & 
        \multicolumn{2}{c}{0.731}  \\
        
        \toprule
    \end{tabular}}
    \vspace{-0.8em}
\label{tab: ab:dataset}
\end{minipage}
\end{minipage}

\textbf{Rejection Sampling}. We conduct a control study by applying the same rejection sampling procedure from our method to the baselines. For preferred data, we select those where the \rlm~can infer the correct decision (called filtered preferred data). This removes possible noisy data. Our method uses this subset, while baselines that require pairwise training data use both this subset and rejected data. As shown in Tab.~\ref{tab: ab:dataset}, our method still achieves the best performance under this setting. By contrast, rejection sampling is not guaranteed to improve the performance of other baselines. These results show that the effectiveness of our method is not due to the higher quality of the training data.\looseness=-1

\textbf{Masked Explicitly Stated Decisions}. We conducted an additional evaluation in which any occurrence of the explicit decision in the generated explanation was masked before passing it to the evaluator model.
As shown in Tab~\ref{tab: masked}, this causes only a slight drop in performance. Specifically, our method’s accuracy decreases by 2.1\% on $\mathtt{SMAC}$, 1.8\% on $\mathtt{MMLU}$, and 3.6\% on $\mathtt{MathQA}$. We find that in most instances where the explanation states the decision explicitly, the decision can still be inferred from the reasoning trace (e.g., detailed numerical derivations or clear contextual cues). In these cases, masking does not significantly hinder the evaluator’s ability to infer the decision. Baselines are also affected under this evaluation protocol, yet our method still outperforms the strongest ones, $\mathtt{KTO}$ and $\mathtt{ReFT}$.\looseness=-1

%the significance of $\mathtt{SFT}$ varies depending on the specific benchmark. 
% Our method achieves the best performance, with an ACC of 78.4\% and an AUC of 0.74 on $\mathtt{MathQA}$. In contrast, 

% while improves to 73.2\% when combined with Majority Voting. This indicates that CoT, when generating reasoning in a single pass, produces more uncertain results. Below, we present a reasoning example from MathQA comparing CoT with our method. The CoT approach employs a complex step-by-step reasoning process, where each decision heavily relies on preceding steps, increasing the likelihood of errors propagating. In contrast, our method offers a more fluent and concise explanation, enabling the Guidance LLM to infer the correct answer more effectively.

\vspace{-0.5em}
\begin{table}[th]
\centering
\caption{A true negative sample from $\mathtt{MMLU}$. Initially, the explanation from the \elm~erroneously supports answer B that focuses on the influence of the filter placement. Post-training with our method, the \elm~accurately justifies the correct answer.}
\vspace{-0.8em}
\begin{tabular}{p{\linewidth}}
\toprule
\textbf{Context} (shortened): A 67-year-old woman had a pulmonary embolism... She underwent placement of an inferior vena cava (IVC) filter. She had a hematoma that was resolving... Which of the following is the most likely cause of this patient's decreased sensation?
    
\hdashrule[0.2ex]{\linewidth}{1pt}{1pt}

\textbf{Decision set} $\bm{\mathcal{A}}$: [A. Cerebral infarction during the hospitalization; B. Complication of the IVC filter placement; C. \textcolor{outputcolor}{Compression} of the lateral femoral cutaneous nerve \textcolor{outputcolor}{$\checkmark$}; D. Hematoma of the left thigh.]

\hdashrule[0.2ex]{\linewidth}{1pt}{1pt}

\textbf{Explanation}: ... The patient's symptoms began after hospitalization, aligning with the placement of the IVC filter... The improvement over time supports a reversible cause related to the {\color{red}filter placement rather than other options... (B,$\boldsymbol{\times}$)}  

\hdashrule[0.2ex]{\linewidth}{1pt}{1pt}

\textbf{Improved Explanation (Ours):} ...involving nerves near sites of cannula placement or previous bleeding sites... The sensation loss around the left thigh aligns with \textcolor{outputcolor}{nerve compression near $\checkmark$}  the sites of recent interventions. \\
\bottomrule
\end{tabular}
\vspace{-1.5em}
\label{tab:example2}
\end{table}

%to improve low-quality explanations
\subsection{Generalization to Negative Samples: Distinguishing True and False Negatives}

As discussed in Sec.~\ref{sec:train}, we train the rectified flow model $\varphi$ solely on positive samples--explanations that allow the \rlm~to infer the ground-truth decision. It is critical that $\varphi$ generalizes to other unseen, negative samples and assigns them reliable rewards. Otherwise, the \elm~would not receive informative reward signals.
These negative samples fall into two categories. True negatives are invalid explanations rightly rejected by the \rlm, while false negatives are effective explanations misclassified by the \rlm. Effective \elm~training requires distinguishing and differentially rewarding these types. We assess whether our method achieves this separation by first using o1-mini to label negatives as true or false, and then examining the rewards that our algorithm assigns to each type.

Fig.~\ref{fig: true_false_negative} demonstrates that our reward model effectively differentiates between the two. Over the course of training, it gradually learns to grant higher rewards to false negatives and lower rewards to true negative samples--precisely the desired behavior. This pattern is mirrored in the accuracy curve: in our framework, rewards are positively related to accuracy, and our reward model achieves approximately 80\% accuracy on false negatives and only around 5\% on true negatives. To see how this improvement happens, Tab.~\ref{tab:example2} gives a true negative example where our method successfully guides the \elm~to refine the explanation, an update attributable to correctly computed rewards. Tab.~\ref{tab:example1} gives an example of a false negative sample. Although the \rlm~cannot infer the correct decision, our method is able to deliver correct rewards. We hypothesize that this generalization capability stems from the cross-attention mechanism. Fig.~\ref{fig: rectified_flow_training} (a) and (b) compare the accuracy of our flow model and the ablation $\mathtt{Ours \ w/o\ Attn}$ on unseen samples. As we can see, without cross-attention, performance on negative samples stagnates throughout training, underscoring its importance. \looseness=-1

\begin{table}[th]
\centering
% \vspace{-0.8em}
\caption{A false negative sample from $\mathtt{MMLU}$ where the explanation is correct but the \rlm~produces an incorrect distribution. Although the rectified flow model was not trained on this sample, it correctly identifies ``the egg roll was present for a substantial time'', a cue missed by the \rlm, and thus provides a correct distribution.}
\vspace{-0.8em}
\begin{tabular}{p{\linewidth}}
\toprule
\textbf{Context} (shortened): A wife and her husband were dining at a restaurant owned by a chef. As the wife walked past a table, she slipped on an egg roll that had been on the floor for quite some time, although the chef was unaware it had fallen there. If she sues the chef 
% for her injuries
, she will most likely:

\hdashrule[0.2ex]{\linewidth}{1pt}{1pt}

\textbf{Decision set} $\bm{\mathcal{A}}$: [A. Recover, because the egg roll on the floor constituted an unsafe condition of the premises; B. Recover, because the egg roll was on the floor \textcolor{outputcolor}{for a substantial period of time before the accident $\checkmark$}; C. Not recover, ...; D. Not recover, ...]

\hdashrule[0.2ex]{\linewidth}{1pt}{1pt}

\textbf{Explanation}: ... The fact that the egg roll was present \textcolor{outputcolor}{for a substantial time $\checkmark$} suggests that the owner should have been aware of the potential hazard and taken steps to address it. This situation falls under premises liability where maintaining safe conditions is crucial to ... % avoid injuries to guests

\hdashrule[0.2ex]{\linewidth}{1pt}{1pt}

\textbf{Distribution $p$ from the \rlm:} [\textbf{0.9297}, 0.0674, 0.0010, 0.0013] $\rightarrow$ `A' {\color{red}$\boldsymbol{\times}$}

\textbf{Distribution $\hat{p}$ from the rectified flow model:} 
[0.0547, \textbf{0.9330}, 0.1089, 0.0685] $\rightarrow$ `B' \textcolor{outputcolor} {$\checkmark$}\\
\bottomrule
\end{tabular}
\vspace{-1.5em}
\label{tab:example1}
\end{table}

\subsection{Ablation Study}
\textbf{Different Backbone Models}. In Tab.~\ref{tab: ft_base}, we change the backbone model of our method to Llama-3.1-8B, where our method still surpasses all baselines, achieving a minimum of 2.6\% higher ACC on $\mathtt{SMAC}$, 2.0\% higher ACC on $\mathtt{MMLU}$, and 3.7\% higher ACC on $\mathtt{MathQA}$. We further evaluate the robustness of our method using two more backbone models: Qwen2.5-7B-Instruct, and Gemma-2-2B-It (abbreviated as Qwen and Gemma).
As shown in Tab~\ref{tab: ab_base_model}, our method consistently improves ACC by up to 19.5\% over $\mathtt{SFT}$, confirming its effectiveness across diverse LLM backbones.

\textbf{Sparse Reward vs Sentence-level Reward}. As shown in Tab.~\ref{tab: ab_arch}, $\mathtt{Ours}$ $\mathtt{(Sp\ Reward)}$ represents the sparse-reward variant and $\mathtt{PPO}$ $\mathtt{(Se\ Reward)}$ represents $\mathtt{PPO}$ with sentence-level reward. Using sparse reward in our method leads to at least 1.2\% lower ACC, indicating that the sentence-level reward design offers better guidance for learning high-quality explanations. With sentence-level reward, $\mathtt{PPO}$ still lags behind $\mathtt{Ours}$ by at least 5.2\% ACC, underscoring the effectiveness of our rectified flow model.\looseness=-1

\begin{minipage}{\textwidth}

\begin{minipage}[t!]{0.60\textwidth}
\makeatletter\def\@captype{table}
        \caption{Accuracy of our method and baselines with a different backbone model, Llama-3.1-8B.}
    \centering
   \vspace{-0.8em}
    \scalebox{0.78}{
    
    \begin{tabular}{crcrcrcrcrcrcrcr}
        \toprule

        \multicolumn{2}{l}{Datasets} &
        \multicolumn{2}{c}{${\mathtt{Ours}}$}&
        \multicolumn{2}{c}{${\mathtt{SFT}}$}&
        \multicolumn{2}{c}{${\mathtt{PPO}}$}&
        \multicolumn{2}{c}{${\mathtt{DPO}}$}&
        \multicolumn{2}{c}{${\mathtt{KTO}}$}&
        \multicolumn{2}{c}{$\mathtt{CoT}$+$\mathtt{SFT}$}&
        \multicolumn{2}{c}{${\mathtt{ReFT}}$}\\

         % \midrule

         \cmidrule(lr){1-2}
        \cmidrule(lr){3-4}
        \cmidrule(lr){5-6}
        \cmidrule(lr){7-8}
           \cmidrule(lr){9-10}
        \cmidrule(lr){11-12}
        \cmidrule(lr){13-14}
        \cmidrule(lr){15-16}

        \multicolumn{2}{l}{${\mathtt{SMAC}}$} & 
        \multicolumn{2}{c}{\textbf{0.689}} &
        \multicolumn{2}{c}{0.583} & 
        \multicolumn{2}{c}{0.632} & 
        \multicolumn{2}{c}{0.650} & 
        \multicolumn{2}{c}{0.655} & 
        \multicolumn{2}{c}{0.614} & 
        \multicolumn{2}{c}{0.663}  \\
        
       \cmidrule(lr){1-2}
        \cmidrule(lr){3-4}
        \cmidrule(lr){5-6}
        \cmidrule(lr){7-8}
           \cmidrule(lr){9-10}
        \cmidrule(lr){11-12}
        \cmidrule(lr){13-14}
        \cmidrule(lr){15-16}

        \multicolumn{2}{l}{${\mathtt{MMLU}}$} & 
        \multicolumn{2}{c}{\textbf{0.724}} &
        \multicolumn{2}{c}{0.638} & 
        \multicolumn{2}{c}{0.682} & 
        \multicolumn{2}{c}{0.698} & 
        \multicolumn{2}{c}{0.673} & 
        \multicolumn{2}{c}{0.653} & 
        \multicolumn{2}{c}{0.704}  \\

        \cmidrule(lr){1-2}
        \cmidrule(lr){3-4}
        \cmidrule(lr){5-6}
        \cmidrule(lr){7-8}
           \cmidrule(lr){9-10}
        \cmidrule(lr){11-12}
        \cmidrule(lr){13-14}
        \cmidrule(lr){15-16}

        \multicolumn{2}{l}{${\mathtt{MathQA}}$} & 
        \multicolumn{2}{c}{\textbf{0.746}} &
        \multicolumn{2}{c}{0.626} & 
        \multicolumn{2}{c}{0.662} & 
        \multicolumn{2}{c}{0.683} & 
        \multicolumn{2}{c}{0.709} & 
        \multicolumn{2}{c}{0.631} & 
        \multicolumn{2}{c}{0.695}  \\

        \toprule
    \end{tabular}}

\label{tab: ft_base}

\end{minipage}
\begin{minipage}[t!]{0.40\textwidth}
\makeatletter\def\@captype{table}
\caption{Accuracy of our method and baselines with masked explicit decisions.}
    \vspace{-0.8em}
    \centering
    \scalebox{0.8}{

    \begin{tabular}{crcrcrcr}
        \toprule

        \multicolumn{2}{l}{Datasets} &
        \multicolumn{2}{c}{${\mathtt{Ours}}$}&
       
        \multicolumn{2}{c}{${\mathtt{KTO}}$}&
        \multicolumn{2}{c}{${\mathtt{ReFT}}$}\\

         % \midrule

         \cmidrule(lr){1-2}
        \cmidrule(lr){3-4}
        \cmidrule(lr){5-6}
        \cmidrule(lr){7-8}

        \multicolumn{2}{l}{${\mathtt{SMAC}}$} & 
        \multicolumn{2}{c}{\textbf{0.743}} & 
        \multicolumn{2}{c}{0.702} & 
        \multicolumn{2}{c}{0.696}  \\
        
       \cmidrule(lr){1-2}
        \cmidrule(lr){3-4}
        \cmidrule(lr){5-6}
        \cmidrule(lr){7-8}

        \multicolumn{2}{l}{${\mathtt{MMLU}}$} & 
        \multicolumn{2}{c}{\textbf{0.754}} & 
        \multicolumn{2}{c}{0.709}  & 
        \multicolumn{2}{c}{0.725}   \\

        \cmidrule(lr){1-2}
        \cmidrule(lr){3-4}
        \cmidrule(lr){5-6}
        \cmidrule(lr){7-8}

        \multicolumn{2}{l}{${\mathtt{MathQA}}$} & 
        \multicolumn{2}{c}{\textbf{0.768} }  & 
        \multicolumn{2}{c}{0.732}  & 
        \multicolumn{2}{c}{0.735}   \\
        
        \toprule
    \end{tabular}}
\label{tab: masked}
\end{minipage}
\end{minipage}

\input{table/ablation}

\textbf{Without Attention}. As shown in Tab.~\ref{tab: ab_arch}, $\mathtt{Ours\ w/o\ Attn}$ underperforms $\mathtt{Ours}$ with a minimum of 5.1\% lower ACC, indicating that the transformer-based architecture can better capture the complex context information. Fig.~\ref{fig: positive} and Fig.~\ref{fig: negative} further elucidate the reason: it cannot even reproduce 100\% accuracy on positive samples and cannot generalize to negative samples. 

%\textbf{Under Noisy Reward}. As shown in Tab.~\ref{tab: ab_arch}, under the noisy reward condition, the performance of our method decreases only slightly from 77.2\% to 74.3\% on unseen tasks in the $\mathtt{MMLU}$ benchmark. Notably, even with these noisy rewards, our method still outperforms $\mathtt{PPO}$ (72.3\%), $\mathtt{Skywork}$ (73.7\%), and $\mathtt{DeepSeek}$ (72.1\%) trained without noise. This demonstrates the robustness of our approach under imperfect reward feedback.

Appendix~\ref{appx:results} reports additional results under noisy reward, varying $\sigma$ and excluding ${\mathtt{SFT}}$.

%suggesting the cross-attention mechanism performs well in the generalization to the negative sample correction.

%consistently improves over base models across different model sizes：不同base model 表格
%attention 作用：mlp， attention 曲线or 表格？

% As illustrated in Fig.~\ref{fig: rectified_flow_training}, the rectified flow model $\varphi$ in $\mathtt{Ours\ w/o\ Attn}$ only has 80\% ACC of identifying positive samples and 20\% ACC of calibrating negative samples.

\section{Conclusion}

We demonstrate that a flow-matching generative model can provide sentence-level and reliable rewards for training LLMs to recover a specific policy and explain that policy's decisions in both RL agents and other LLMs. Looking ahead, we envision extending this approach into a general-purpose LLM training framework that automatically generates high-quality reward signals.

\section{Ethics Statement}
To assess the quality of the synthetic dataset generated by o1-mini, we enlist 12 participants (2 females, 10 males, aged 20-40) to evaluate whether the samples are reasonable and human-like. For the human evaluation, we recruited 45 participants (30 male, 15 female, aged 20–40) from diverse fields, including engineering, chemistry, machine learning, and economics. To ensure consistency, we provided standardized instructions and illustrative examples before the evaluation. All participant information was anonymized, and no personally identifiable data were collected, ensuring that there was no risk to privacy.

\section{Reproducibility Statement}
We have undertaken comprehensive efforts to ensure the reproducibility of this work. The complete source code is provided in the supplementary materials. For theoretical analysis, detailed explanations of underlying assumptions and full proofs of the claims are presented in Appx.~\ref{appx:theorem}. A complete and transparent description of the data processing steps employed in our experiments is also included in the supplementary materials.

% Our method has the potential to facilitate human-AI collaboration applications, such as transportation, education, and security defense.

%\section{Impact Statements}
%This paper presents work whose goal is to advance the field of machine learning by developing a model-agnostic explanation generator for intelligent agents, enhancing transparency and interpretability in agent decision prediction. The ability to generate effective and interpretable explanations has the potential to foster trust in AI systems, improving effectiveness in high-stakes applications such as healthcare, finance, and autonomous systems. Overall, we believe our work contributes positively to the broader AI ecosystem by promoting more explainable and trustworthy AI.

\bibliography{iclr2026_conference}
\bibliographystyle{iclr2026_conference}

\newpage
\appendix

\section{More Related Works}\label{appx:related_work}

\textbf{Diffusion in Transformer} (DiT,~\citep{yang2023diffusion}) leverages the strengths of self-attention of Transformers to improve the performance of diffusion models across a range of tasks, including image and text generation~\citep{cao2024survey}. \citep{dhariwal2021diffusion} demonstrate how Transformer-based architectures can optimize the denoising process in diffusion models, resulting in high-quality image synthesis. \citep{ulhaq2022efficient} explore efficient implementations for diffusion within Transformer. These works are related to our work, as we embed flow matching into the last layer of an LLM. The difference is that we train the LLM-augmented flow matching model as a reward model for the purpose of LLM post-training.

\textbf{Cross-attention} is a popular technique for processing information across multiple modalities~\citep{radford2021learning, alayrac2022flamingo, li2023blip}. Approaches such as T2I-Adapter~\citep{mou2024t2i} and VMix~\citep{wu2024vmix} use cross-attention mechanisms between text encoders (an LLM) and diffusion models to enhance the generation of high-quality images from textual descriptions. More generally, cross-attention has helped solve tasks that require both vision and language understanding~\citep{hatamizadeh2025diffit, cao2024survey}.  Different from previous work on DiT and cross-attention-based image/video generation, to our best knowledge, the proposed method is the first to use generative models and cross-attention to generate rewards for RL-based LLM training.

\textbf{Explainable AL/RL without natural language explanations}. Our method is suited within the domain of explainable AI~\citep{arrieta2020explainable,carvalho2019machine,ehsan2019automated,gunning2017explainable,ras2018explanation,gilpin2018explaining}. \emph{Faithfulness}~\citep{jacovi2020towards} is one of key concepts from the XAI literature, which is defined as the degree to which an explanation reflects the model’s actual decision-making process, rather than merely presenting a narrative that sounds plausible to humans (i.e., plausibility).

Recent studies have explored how to generate faithful explanations from large language models. Some approaches~\citep{lyu2023faithful,radhakrishnan2023question} focus on enhancing faithfulness through Chain-of-Thought prompting, breaking down the reasoning process into multiple sub-sentences. Others~\citep{schimanski2024towards} improve faithfulness by augmenting the training data to provide stronger supervisory signals. 
For measuring rational faithfulness, \cite{matton2025walk} introduces a causal counterfactual evaluation framework using an auxiliary LLM and a Bayesian hierarchical model. \cite{parcalabescu2023measuring} proposes CC-SHAP, a token-level self-consistency measure that records model contributions. \cite{madsen2024self} measures self-explanations and find that faithfulness varies significantly across models and datasets. Our method captures the essential behavior of the model by faithfully reproducing its outputs from the given inputs, thereby serving as an interpretable intermediate representation of model behavior. In this sense, our approach offers a form of faithfulness from a functional and behavioral perspective.

% Our work takes a different yet complementary path: we optimize explanations based on how well they support accurate inference of the model’s decision. Our approach can be combined with both reasoning decomposition and data augmentation techniques for further improvements. In contrast, our approach introduces a probabilistic measurement framework, which captures the inherently subjective nature of explanations. The reward distribution generated by our method encourages the Explanation LLM to generate explanations that genuinely support the inference of the ground-truth decision by a variety of users or agents—rather than simply appearing plausible to a single individual. 

For explainable RL (XRL), our approach draws particular parallels with it. \emph{Post-hoc XRL} methods focus on relating inputs and outputs of a trained RL policy in an interpretable way, using an interpretable \emph{surrogate} model as policy approximation. Examples of surrogate models include imitation learning~\citep{abbeel2004apprenticeship}, learning from demonstration~\citep{argall2009survey}, finite state machines~\citep{koul2018learning, danesh2021re}, decision trees and their variants~\citep{bastani2018verifiable,jhunjhunwala2019policy,bewley2021tripletree,liu2018toward}, genetic programming~\citep{zhang2020interpretable}, programmatic policy searching~\citep{verma2018programmatically}, and deterministic finite automata~\citep{hasanbeig2021deepsynth}. However, in order to be interpretable, surrogate models are designed as simple as possible.
\emph{Ad-hoc XRL} methods represent policies as inherently interpretable models. For example, \cite{silva2019optimization,topin2019generation,hein2018interpretable, landajuela2021discovering} use decision trees as policy approximators. However, the capacity of these models is typically limited. Saliency maps distinguish observation elements that influence decisions~\citep{atrey2019exploratory,greydanus2018visualizing,gottesman2020interpretable}, but does not capture the reasoning behind decisions~\citep{atrey2019exploratory}, leaving humans to give ad-hoc explanations based on these visual cues. 
Example-based methods use similar experiences to explain decisions~\citep{amir2018highlights,huang2018establishing,zahavy2016graying,topin2019generation}, but largely overlook underlying reasoning.

\textbf{Use explanations to train LLMs}. Various prior works have explored training or tuning language models with explanatory cues, often relying on techniques such as span or word highlighting~\citep{hase2021can,zhou2020towards,narang2020wt5,rajani2019explain}. Moreover, explicitly training language models with natural language explanations or instructions has been shown to enhance their ability to leverage explanations in prompts~\citep{wei2021finetuned}. These works are different from our method, because we do not use explanations to train LLMs. Instead, we train LLMs to generate explanations.

\section{Notations for Transformers}\label{appx:preliminary}

We consider an LLM with $L$ layers, the hidden state after $l$ layers, $\idx[\mH][][n,l]$, is projected by three weight matrices \(\mW_Q \), \(\mW_K \), and \(\mW_V \) to the query, key, and value embeddings $\idx[\mQ][][n,l]$, $\idx[\mK][][n,l]$, and $\idx[\mV][][n,l]$, respectively. The self-attention is calculated as:
\begin{align}
(\idx[\mQ][][n,l], \idx[\mK][][n,l], \idx[\mV][][n,l]) &= \idx[\mH][][n,l] (\mW_Q, \mW_K, \mW_V), \nonumber\\
\idx[\mA][][n,l] &= \idx[\mQ][][n,l] {\idx[\mK][][n,l]}^\top/\sqrt{d_K},\nonumber\\
\text{Attn}(\idx[\mH][][n,l]) &= \sigma(\idx[\mA][][n,l]) \idx[\mV][][n,l],\nonumber
\end{align}
where $\sigma(\cdot)$ is SoftMax, and $\mA$ is the self-attention matrix. 

\section{Learning Framework}\label{appx:methods}

\textbf{\elm}. Our method is developed around the \elm. Given a set $\mathcal{D}_e=\{(a_j,c_j)\}_{j=1}^{J}$ of decisions $a_j$ and their context $c_j$, we use the following prompt $x^e_j$ to ask the \elm~to generate explanations: $\mathtt{Given}$ $\mathtt{\left[Context\ c_j\right].}$ $\mathtt{Please}$ $\mathtt{analyze}$ $\mathtt{reasoning}$ $\mathtt{for}$ $\mathtt{the}$ $\mathtt{agent}$ $\mathtt{decision}$ $\mathtt{based}$ $\mathtt{on}$ $\mathtt{the}$ $\mathtt{context}$. 

Suppose that the \elm~generates $K_e$ sentences as output: $y^e_j\shorte (s_j^{e(1)}, \cdots,s_j^{e(K_e)}])$. For the generated content to be effective explanations, we hope that one can consistently infer agent decisions from explanations across various contexts. To this end, we seek feedback regarding how likely the actual decision $a_j$ is a plausible outcome given each incremental portion of the explanation. 

Such feedback is most accurate when provided by human annotators. However, human feedback is expensive~\citep{bai2022constitutional}. We first discuss an alternative approach that uses a \rlm~as a surrogate feedback provider, whose downside will motivate the proposed generative reward method introduced in the next subsection. 

We query the \rlm~with the prompt: $\mathtt{Given}$ $\mathtt{[Context\ c_j],}$ $\mathtt{the\ reasoning\ is\ }$ $s_j^{e(1:k)}$ $\mathtt{].}$ $\mathtt{Thus,}$ $\mathtt{the}$ $\mathtt{decision\ is\ [a\ decision}$ $a\in\mathcal{A}\mathtt{]}$. Here, $s_j^{e(1:k)}$ is the first $k$ sentences of the explanation $y^e_j$. We denote this prompt to the \rlm~by $x_j^{g(k)}(y_j^e)$. The dependence on $y_j^e$ will be omitted when unambiguous. 

We are interested in the likelihood a decision $a\in\mathcal{A}$ appearing at the end of $x_j^{g(k)}(y_j^e)$, influenced by logits $M(a | x_j^{g(k)}(y_j^e))$ (Eq.~\ref{equ:token_logits}). In practice, $a$ is represented by some tokens describing the decision. If it involves multiple tokens, we calculate their mean~\citep{yang2022logit}.

By applying the SoftMax operation to $M(a | x_j^{g(k)}(y_j^e))$, $a\in\mathcal{A}$, we get a distribution over decisions:
\begin{align}
    p(a_j | x_j^{g(k)}) = \textsc{SoftMax}_j\left(M(a | x_j^{g(k)})\right).
\end{align}

This distribution is defined for the first $k$ sentences, $k=1,\cdots,K_e$. Intuitively, the distribution $p$ changes as we feed the sentences in the explanation $y_j^e$ incrementally. These changes measure the contribution of each sentence to the effectiveness of the explanation, allowing us to define per-sentence rewards for $y_j^e$ as the changes of the likelihood of the actual decision $a_j$ after each newly added sentence:
\begin{align}
     r(s_j^{e(k)}) = p(a_j | x_j^{g(k)}) - p(a_j | x_j^{g(k-1)}),\label{equ:elm_reward}
\end{align}
which can be understood as an information gain~\citep{ton2024understanding}. Calculating sentence-level rewards is a trade-off~\citep{deepseekai2025deepseekr1incentivizingreasoningcapability}. We benefit from denser reward signals compared to a single reward for the whole explanation, and also avoid the costs of per-token reward calculation.

We intentionally exclude the actual decision $a_j$ from the prompt $x_e$ to the \elm. Otherwise, the rewards $r(s_j^{e(k)})$ might be trivial and encourage merely restating the decision: the sentence that discloses the decision will get a very large reward, while the following sentences get fairly small rewards, regardless of their content.

The disadvantage of directly using this \rlm~is that the rewards $r(s_j^{e(k)})$ could be noisy or inefficient~\citep{yang2024regularizing}, as proven by a baseline method $\mathtt{Proxy}$ $\mathtt{LLM}$ in Tab.~\ref{tab: ft_instruct}. We propose to fix this problem by introducing a rectified flow model for reward generation.

\textbf{Rectified Flow Model Architecture}.

\emph{Cross-Attention}. $(W_Q, W_K, W_V)$ are last‐layer attention weights of the \elm. We apply these weights to the flow-token hidden states $(H_{EMB})$ to obtain corresponding queries, keys, and values: $(Q_{EMB}, K_{EMB}, V_{EMB}) = H_{EMB} (W_Q, W_K, W_V)$. In parallel, we also obtain the last-layer queries, keys, and values from the \elm~for the given context (i.e., the query and its explanation), denoted as $Q^{(L-1)}, K^{(L-1)}, V^{(L-1)}$, where L is the number of transformer layers. We then perform standard multi-head attention using the concatenated queries, keys, and values $([Q^{(L-1)}, Q_{EMB}], [K^{(L-1)}, K_{EMB}], [V^{(L-1)}, V_{EMB}])$. Finally, we extract the attention hidden states at the flow-token positions as the output.

\emph{Projector}. Define $\vh_{\textsc{Attn},\vz_t}$ as the latent state of the flow token $\vh_{\textsc{Emb},\vz_t}$ after cross-attention. This state has incorporated the ODE time $t$ and explanatory information through cross-attention. Progressing from this state, we use a four-layer fully-connected network $\varphi_{\textsc{Proj}}:\mathbb{R}^d\rightarrow\mathbb{R}^{|\mathcal{A}|}$ with ReLU activation and layer normalization to generate the vector field $\varphi(t,\vz_t) = \varphi_{\textsc{Proj}}(\vh_{\textsc{Attn},\vz_t})$. We find that skip-layer connections are important for training stability. Specifically, we append the inputs $\vz_t$ and $t$ to the hidden layers of $\varphi_{\textsc{Proj}}$.

\section{Rectified Flow can Correct Errors}\label{appx:theorem}

\begin{figure*}[t]
    \centering
    % \vspace{-0.5em}
    \includegraphics[width=\linewidth]{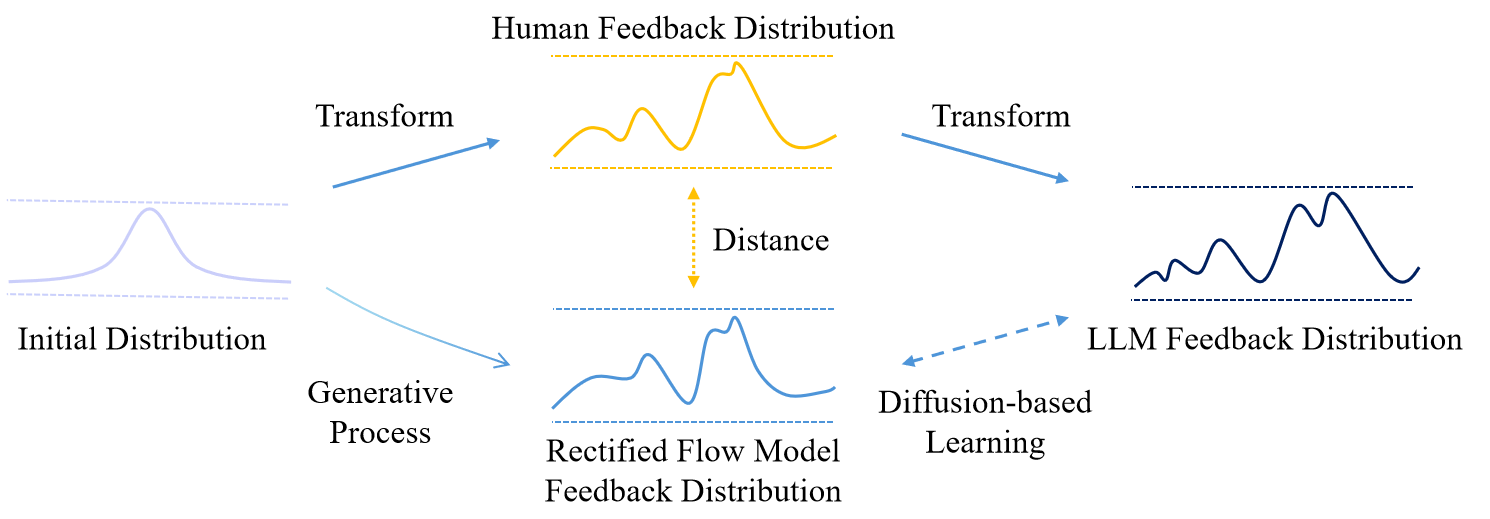}
    % \vspace{-1.5em}
    \caption{Overview of the idea of our theoretical analysis for the proposed flow-matching–based reward generation framework. We train a flow-matching model to approximate the distribution of LLM-generated feedback. Assuming the LLM feedback distribution is a transformation of the ground-truth human feedback distribution, we derive a closed-form expression for the distance between the induced flow-matching reward distribution and the human feedback distribution.}
    \vspace{-0.5em}
    \label{fig:diffusion}
\end{figure*}

\flowerror*

\begin{proof}
Recall that for any two probability measures $\mu,\nu$ on $\mathbb{R}^d$, the quadratic Wasserstein distance is

$$
W_{2}^{2}(\mu,\nu)
=
\inf_{\gamma\in\Gamma(\mu,\nu)}
\int_{\mathbb{R}^d\times\mathbb{R}^d}\|x-y\|^2\,d\gamma(x,y),
$$

where $\Gamma(\mu,\nu)$ is the set of all couplings of $\mu$ and $\nu$.

Moreover, if $f:\mathbb{R}^d\to\mathbb{R}^d$ is $L$-Lipschitz then

$$
W_{2}\bigl(f_{\sharp}\mu,\;f_{\sharp}\nu\bigr)
\;\le\;
L\,W_{2}(\mu,\nu).
$$

We want to bound
$\;W_{2}\bigl(p_{\textsc{flow}},\,p\bigr)$.
By the triangle inequality,

$$
W_{2}(p_{\textsc{flow}},\,p)
\;\le\;
W_{2}\bigl(p_{\textsc{flow}},\,q_1\bigr)
\;+\;
W_{2}\bigl(q_1,\;\hat p\bigr)
\;+\;
W_{2}\bigl(\hat p,\;p\bigr).
$$

The middle term is exactly the training error:
  $\;W_{2}(q_1,\hat p)\le\varepsilon.$
For the last term, $W_{2}(\hat p,p)$, we observe the fact that $\hat p=p*p_{0}$ shares its mean with $p$; since the corruption is mean-zero Gaussian, the optimal coupling between $p$ and $\hat p$ is to pair each clean sample $x$ with $x\!+\!\xi$ where $\xi\sim p_{0}$, giving zero squared cost in expectation.

Hence it suffices to bound the first term,
$\;W_{2}(p_{\textsc{flow}},\,q_1)$.

Since

$$
p_{\textsc{flow}}
\;=\;
\phi_{1\sharp}p_{0},
\qquad
q_{1}
\;=\;
\phi_{1\sharp}q_{0},
$$

and $\phi_{1}$ is $L$-Lipschitz by construction,

$$
W_{2}\bigl(p_{\textsc{flow}},\,q_{1}\bigr)
=
W_{2}\bigl(\phi_{1\sharp}p_{0},\,\phi_{1\sharp}q_{0}\bigr)
\;\le\;
L\,W_{2}\bigl(p_{0},\,q_{0}\bigr).
$$

For two zero-mean Gaussians in $\mathbb{R}^{|\mathcal A|}$,
$\mathcal N(0,\sigma_r^2I)$ and $\mathcal N(0,\sigma^2I)$, one shows

$$
W_{2}\bigl(p_{0},\,q_{0}\bigr)
=
\sqrt{|\mathcal A|}\,\bigl|\sigma_{r}-\sigma\bigr|.
$$

Combining,

$$
W_{2}(p_{\textsc{flow}},\,p)
\;\le\;
L\,\sqrt{|\mathcal A|}\,\bigl|\sigma-\sigma_{r}\bigr|
\;+\;
\varepsilon,
$$

which is exactly the claimed bound.
\end{proof}

\error*

\begin{proof}
First, by the definition of \(\varepsilon'\) as the model’s approximation error to \(\hat p\), we have
\[
W_{2}(\tilde p,\;\hat p)\;\le\;\varepsilon'.
\]
Next, apply the triangle inequality for the \(2\)-Wasserstein distance:
\[
W_{2}(\tilde p,\,p)
\;\le\;
W_{2}(\tilde p,\,\hat p)
\;+\;
W_{2}(\hat p,\,p).
\]
Combining with the above bound gives
\[
W_{2}(\tilde p,\,p)
\;\le\;
\varepsilon'
\;+\;
W_{2}(\hat p,\,p).
\]
It remains only to compute \(W_{2}(\hat p,p)\).  Recall that
\(\hat p = p * p_{0}\) where \(p_{0} = \mathcal N(0,\sigma_r^2 I_{|\mathcal A|})\).  Since \(p\) and \(\hat p\) differ by the addition of a zero-mean Gaussian of covariance \(\sigma_r^2 I_{|\mathcal A|}\), one shows directly for two zero-mean Gaussians in \(\mathbb{R}^{|\mathcal A|}\) that
\[
W_{2}\bigl(\hat p,\,p\bigr)
=
\sqrt{|\mathcal A|}\,\bigl|\sigma_r\bigr|.
\]
Hence
\[
W_{2}(\tilde p,\,p)
\;\le\;
\varepsilon'
\;+\;
\sqrt{|\mathcal A|}\,\bigl|\sigma_r\bigr|,
\]
as claimed.
\end{proof}

\subsection{With Different Functional Form}

The following is the general case where the initial distribution $q_{0}$ and the error model $p_{0}$ follow different laws.

For the true distribution $p$ and the flow model’s output $p_{\text{flow}}$, the error is bounded by

$$ W_{2}(p_{\text{flow}},p)\le \epsilon +\sqrt{\operatorname{tr}\Sigma_{p_{0}}} +L\sqrt{|\mu_{p_0}|^2 +\operatorname{tr}\Sigma_{p_0} +\operatorname{tr}\Sigma_{q_0}}, $$

where $\Sigma_{\cdot}$ denotes the covariance matrix and $\mu_{\cdot}$ is the mean. When both $p_{0}$ and $q_{0}$ are Gaussian, this reduces exactly to Theorem 1.

The human‑feedback distribution is unknown, but we assume we observe a noisy proxy provided by the LLM with an additive noise model $p_0$. In the proof, the flow network acts as a denoiser: it transforms the proxy toward the latent human distribution. The inequality above quantifies how closely the denoised distribution can approximate the true one.

\section{Prompts}\label{appx:prompt}
Below are the prompts used for the \elm~and the \rlm~on the $\mathtt{SMAC}$ datasets. For the \elm, the prompt describes the overview of the scenarios along with the previous action-state pairs and instructs the \elm~to analyze the provided information. In response, the \elm~generates a summary of the preceding trajectories and implies the action selection. For the \rlm, the prompt includes a statement summarizing both the input to and output from the \elm, followed by the phrase: \emph{'Therefore, the missing action is {a possible action from $\mathcal{A}$}'}. Then, we calculate the mean token logits for each possible action in $\mathcal{A}$, which are then used to determine the action with the highest likelihood.

\begin{tcolorbox}[title=An example that the prompts used for the \elm~and the \rlm~on the $\mathtt{SMAC}$ datasets]
    \textbf{\textcolor{usercolor}{User for \elm:}} \\
   Q: Here are 5 allies and 6 enemies in $\mathtt{SMAC}$. The goal of each ally is to attack all the enemies together with other allies. At every step, each ally can choose one action to take from the Action set. The trajectory of each ally consists of a series of {State, Action, Value} pairs. Here are 4 consecutive {State, Action, Value} pairs where the last state-action pair misses the Action: xxx. Please help me concisely summarize and analyze this part of the trajectory for the missing Action prediction.\\
    
    \vspace{1ex}
    \textbf{\textcolor{cotcolor}{\elm:}} \\
R: Throughout the provided trajectory, the ally consistently maneuvered strategically while maintaining full health and ensuring weapon readiness. Initially, moving SOUTH likely aimed to xxx. The subsequent WEST movement improved xxx. \textbf{The predicted next action should be a directional movement aimed at advancing toward a more optimal attacking position. Compared to other directions, EAST offers the best alignment for coordinated attacks, ensuring allies can effectively converge on and target enemies together.}\\
    
    \vspace{1ex}
    \textbf{\textcolor{outputcolor}{User for \rlm:}} \\
Q + R + Therefore, the missing action is \{a possible action from $\mathcal{A}$\}.\\
 
 \vspace{1ex}
 \textbf{\textcolor{outputcolor}{\rlm:}}\\
The token logits of the given possible action.\\

\label{box: flow_smac}
\end{tcolorbox}

\begin{tcolorbox}[title=An example that the prompts used for the \elm~and the \rlm~on the $\mathtt{AI2}$-$\mathtt{THOR}$ datasets]
    \textbf{\textcolor{usercolor}{User for \elm:}} \\
   Q: There are multiple robots in the simulation,each with different skills and capabilities. The robots are tasked with performing a series of actions in a simulated environment to complete a certain task. Robot one xxx. Task is xxx. Current environment condition is xxx. The previous actions include xxx. The available actions are xxx. You act as an AI assistant to help the robots decompose the tasks into actions step by step.\\
    
    \vspace{1ex}
    \textbf{\textcolor{cotcolor}{\elm:}} \\
R: Toggling on the television and placing the keychain on the coffee table. Robot 0 now should take a corrective navigation step \textbf{(GoToObject)} before successfully executing ToggleOnObject. After completing this sub-task, Robot 1 should navigate to the coffee table and keychain, pick up the keychain, and place it on the table in the correct order.\\
    
    \vspace{1ex}
    \textbf{\textcolor{outputcolor}{User for \rlm:}} \\
Q + R + Therefore, the next action is \{a possible action from $\mathcal{A}$\}.\\
 
 \vspace{1ex}
 \textbf{\textcolor{outputcolor}{\rlm:}}\\
The token logits of the given possible action.\\

\label{box: flow_ai2thor}
\end{tcolorbox}

Above are the prompts used for the \elm~and the \rlm~on the $\mathtt{AI2}$-$\mathtt{THOR}$ datasets. For the \elm, the prompt describes the overview of the scenarios along with the previous action pairs and instructs the \elm~to help robots to take action step by step. In response, the \elm~generates potential trajectories and implies the action selection. For the \rlm, the prompt includes a statement summarizing both the input to and output from the \elm, followed by the phrase: \emph{'Therefore, the next action is {a possible action from $\mathcal{A}$}'}. Then, we calculate the mean token logits for each possible action in $\mathcal{A}$, which are then used to determine the action with the highest likelihood.

\begin{tcolorbox}[title=An example that the prompts used for the \elm~and the \rlm~on the $\mathtt{MMLU}$ datasets]
    \textbf{\textcolor{usercolor}{User for \elm:}} \\
   Q: Below is a multiple-choice question about professional law. Please explain your reasoning without explicitly revealing your selection. The question is: Which of the following is not within the power of a federal subpoena to command a person to do at a specified time and place? A: Submit to a physical or mental examination.; B: Attend and testify as to persons and/or events related to a cause of action.; C: Produce designated documents, electronically stored information, or tangible things in that person's possession, custody, or control.; D: Permit the inspection of premises.\\
    
    \vspace{1ex}
    \textbf{\textcolor{cotcolor}{\elm:}} \\
R: Federal subpoenas are designed to compel testimony, the production of documents, and permit inspections related to a case. However, requiring an individual to undergo a physical or mental examination typically falls outside the scope of what a subpoena can mandate. \textbf{Such examinations usually require separate legal processes or warrants, ensuring individuals' rights are adequately protected.}\\
    
    \vspace{1ex}
    \textbf{\textcolor{outputcolor}{User for \rlm:}} \\
Q + R + Therefore, the correct answer is \{a possible answer from $\mathcal{A}$\}.\\
 
 \vspace{1ex}
 \textbf{\textcolor{outputcolor}{\rlm:}}\\
The token logits of the given possible answer.\\
\label{box: flow_mmlu}
\end{tcolorbox}

\begin{tcolorbox}[title=An example that the prompts used for the \elm~and the \rlm~on the $\mathtt{MathQA}$ datasets]
    \textbf{\textcolor{usercolor}{User for \elm:}} \\
   Q: Below is a multiple-choice question about math problem. Please explain your reasoning without explicitly revealing your selection. The question is: an auction house charges a commission of 18 \% on the first \$50,000 of the sale price of an item, plus 10 \% on the amount of the sale price in excess of \$50,000. what was the price of a painting for which the house charged a total commission of \$24,000? a ) \$115,000, b ) \$160,000, c ) \$ 200,000, d ) \$ 240,000, e ) \$365,000\\
    
    \vspace{1ex}
    \textbf{\textcolor{cotcolor}{\elm:}} \\
R: \textbf{Say the price of the house was \$ x, then 0.18 * 50,000 + 0.1 * ( x - 50,000 ) = 24,000 $\rightarrow$ x = \$ 200,000} ( 18 \% of \$ 50,000 plus 10 \% of the amount in excess of \$ 50,000, which is x - 50,000, should equal to total commission of \$ 24,000 )\\
    
    \vspace{1ex}
    \textbf{\textcolor{outputcolor}{User for \rlm:}} \\
Q + R + Therefore, the correct answer is \{a possible answer from $\mathcal{A}$\}.\\
 
 \vspace{1ex}
 \textbf{\textcolor{outputcolor}{\rlm:}}\\
The token logits of the given possible answer.\\
\label{box: flow_mathqa}
\end{tcolorbox}

There are the prompts used for the \elm~and the \rlm~on the $\mathtt{MMLU}$ and $\mathtt{MathQA}$ datasets. For the \elm, the user provides the question along with the possible answer options and requests the \elm~to explain the reasoning process without explicitly revealing the selection. In response, the \elm generates an explanation that analyzes the problem and implicitly suggests an answer. For the \rlm, the user presents a statement that includes both the input to and the output from the \elm, followed by the phrase: \emph{'Therefore, the correct answer is {a possible answer from $\mathcal{A}$}'}. to calculate the mean token logits of the given possible answer. Then, we calculate the mean token logits for each possible answer in $\mathcal{A}$, which are then used to determine the answer with the highest likelihood.

\section{Additional Experimental Setup}\label{appx:experimental_setup}

\textbf{Benchmark}. 

\textbf{(1) $\mathtt{SMAC}$}. We generate explanations for an ally agent based on action-state history. Our dataset consists of 2K trajectories (1.5K for training and 0.5K for evaluation), with each trajectory containing states, actions, and rewards for 30 time steps. We feed information of previous 4 consecutive time steps to the \elm~to generate explanations.

Specifically, we first train an RL policy using MAPPO~\citep{yu2022surprising} on $\mathtt{SMAC}$ and then collect the action-state pairs at each timestep during the evaluation phase. Each action-state pair contains information about the current agent's position, health, and weapon cooldown, as well as the corresponding details for visible allies and enemies. Additionally, the action selected by the current agent is included. All action-state pairs are stored in JSON format:

\emph{$\begin{aligned}State:
&\left\{visible \: allies: \left\{relative \left(x,y\right); weapon\_cooldown; health\right\}\right.\\
&\left\{id 0:\left[ 0.0765, -0.0765\right]; 0.0; 1.0\right\}; \left\{id 1:\left[0.0765, 0.0\right]; 0.0; 1.0\right\}; \\
&\left\{id 2:\left[0.0765, 0.0765\right]; 0.0; 1.0\right\}; \left\{id 3:\left[0.153, 0.0\right]; 0.0; 1.0\right\}; \\
&None \: visible \: enemies; own\_health: 1.0\left.\right\}; Action: SOUTH
\end{aligned}$}

We feed the \elm~with 4 consecutive action-state pairs, masking the action in the final pair, and then instruct the \elm to analyze the provided information and infer the missing action.

\textbf{(2) $\mathtt{AI2}$-$\mathtt{THOR}$}. We evaluate our method on AI2-THOR, an embodied AI benchmark widely used in RL research, thereby testing the proposed method in a fully sequential decision-making environment. 

Specifically, AI2-THOR includes household activity scenarios that require multi-step planning, multi-robot coordination, and interaction with dynamic environments. We use the \emph{Complex Tasks} split from the SMART-LLM dataset, in which each task involves 2–4 heterogeneous robots with complementary skills. These tasks are challenging because they cannot be decomposed into independent sub-problems solvable by a single robot; agents must reason over past actions and observations, perform long-horizon planning, and coordinate multiple robots strategically to leverage their combined capabilities. This setting demands strong reasoning and sequencial decision-making, making it a suitable benchmark for evaluating our method's ability to recover a complex underlying policy. The original Complex Tasks split contains only 8 tasks. To broaden task diversity, we expand the set to 59 tasks by prompting GPT-4o to generate additional tasks that follow the same structural requirements. 

The prompt is formatted as: \\
\emph{\\ AI2THOR is a embodied-AI benchmark xxx.\\
The robots should be xxx.\\
The tasks should be xxx.\\
Here are some examples of embodied-AI task.\\
Example 1: xxx. \\
...\\
Example n: xxx.\\
Please help me generate new tasks that satisfy the above requirement.}\\
After generating these new tasks, we ask both human and GPT-4o to verify the correctness of these tasks.

We then apply RL to obtain a near-optimal policy over these 59 tasks and record its successful trajectories of 70\% tasks for training our method and baselines. During evaluation, we measure the average success rate of our method and baselines on the remaining 30\% unseen tasks to assess their generalization capability.

\textbf{(3) $\mathtt{MMLU}$}. We choose 4 challenging Professional Knowledge subsets (Professional214
Medicine (272 samples), Professional Law (1.53K samples), Professional Accounting (282 samples),215
and Professional Psychology (612 samples)). For each dataset, 70\% of the samples are selected randomly for training, and the remaining 30\% for evaluation.

\textbf{Training Details}. For training dataset construction, established instruction-tuning datasets provide (general-purpose) preference data for answers, but we optimize explanations for answers. As a result, existing instruction-tuning datasets can’t perform very well in our tasks. Constructing of a human-annotated dataset is labor-intensive, so we leverage OpenAI o1-mini to generate explanations for each sample, providing a more efficient alternative.
To generate positive samples, o1-mini is prompted with a question and its correct answer. The prompt used for positive sample generation is as follows: \emph{'Below is a multiple-choice question about xxx. Please explain your reasoning without explicitly revealing the correct answer: xxx. The question is: xxx . The options are xxx.'}.
Similarly, to generate negative samples, o1-mini receives a question with a randomly selected incorrect answer in the prompt: \emph{'Below is a multiple-choice question about xxx. Please explain your reasoning without explicitly revealing a wrong answer: xxx. The question is: xxx . The options are xxx.'}.

To further assess the quality of the generated samples, we enlist 12 participants (2 females, 10 males, aged 20-40) to evaluate whether the samples are reasonable and human-like. We randomly select 5\% randomly chosen samples from each datasets and ask panticipants to rate them on a scale of 1 to 4: 1 = unreadable, 2 = readable but unreasonable, 3 = reasonable but unnatural, and 4 = natural and human-like. The example of a sample is listed below.

\begin{tcolorbox}[title=An example that the evaluation of the quality of our constructed dataset on $\mathtt{MathQA}$]

    \textbf{Q} = Below is a multiple-choice question about professional law. Please explain your reasoning without explicitly revealing your selection. The question is: At a defendant's trial for drug dealing, a prosecution witness testified that he had heard the defendant telling a group of people to come to the defendant's house because he had "a great crop of apples for sale." The prosecutor then called and qualified a drug enforcement agent as an expert in how drug dealing is conducted. The prosecutor now seeks to have the expert testify that in her opinion, based on years of experience with drug dealers, the defendant's statement about the apples was code for his having drugs for sale. Should the expert's opinion be admitted? \\
    
    \textbf{Decision set $\mathcal{A}$} = 
    
[\textbf{a}: No, because an expert must base his or her opinion on scientific or technical data, not merely on experience.; 
    
\textbf{b}: No, because the opinion is based on the criminal actions of others engaged in drug dealing, and thus is substantially more unfairly prejudicial than probative.;

\textbf{c}: Yes, because the opinion is based on her specialized knowledge, and it will assist the trier of fact in understanding the evidence and determining facts in issue in the case.; 

\textbf{d}: Yes, provided that the expert first testifies about the specific information upon which she based her view that the defendant was using "apples" as a code word to refer to drugs.]\\
    % In the provided trajectory, each ally demonstrates a cautious yet proactive approach to engaging enemies. 
    
    \textbf{{Explanation:}} The expert’s testimony is likely admissible because her specialized knowledge can clarify the defendant’s statements in the context of drug dealing. Courts generally allow experts to provide opinions that assist the jury in understanding evidence or determining facts. As long as her opinion is based on her experience and doesn’t rely solely on hearsay, it should help the trier of fact make informed decisions.  \\
    
    Score: (1-4, 1=unreadable, 2=readable but unreasonable, 3=reasonable but unnatural, 4= natural and human-like):\\\

\end{tcolorbox}

We train the \elm~and the rectified flow model iteratively. In the first step, \elm~generates an explanation for each sample 3 times in the training set. The \rlm~then classifies these explanations as positive or negative. We use the positive samples to train the rectified flow model. Once trained, the rectified flow model is employed in subsequent the \elm~training. This entire process constitutes one round. As shown in Fig.~\ref{fig: elm_training}, the \elm’s accuracy plateaus during the third round because the updates to the reward model no longer yield significant performance improvements. By comparison, in the first two rounds, the accuracy initially increases rapidly due to substantial updates to the reward model, followed by a more gradual improvement. Therefore, we perform two rounds of training. Note that the training process can be adapted for online learning, where the \elm~and the rectified flow model are trained simultaneously.

In each round, the \elm and the flow model are trained for 10 and 100 epochs, respectively. The learning rate for the \elm is initialized as $2e\shortn 5$ and then decays linearly to zero, whereas the learning rate for the flow model is fixed at $2e\shortn 4$. \elm~uses the Adam optimizer, BF16 and LoRA with hyperparameters $r_{lora}=16$ and $\alpha_{lora}=16$, while DeepSpeed ZeRO-3 is employed to accelerate training. The rectified flow model uses the Adam optimizer. 
The batch size is 12 for the \elm~training and 128 for the flow model training. 
We run our method on 4 parallel 80GB Nvidia A100 GPUs.
For \elm~training, each GPU processes at most 1 sample at a time, with gradient accumulation over 3 steps. For the flow model training, each GPU processes at most 32 samples at a time. ${\mathtt{SFT}}$ typically takes 0.5 to 1 hour, while our method requires approximately 14--30 hours on different datasets.

\textbf{Evaluation}. For evaluation, we use temperature $\tau=0.7$ across all settings. The maximum number of generated tokens is 350 for $\mathtt{SMAC}$, and 200 for $\mathtt{MMLU}$ and $\mathtt{MathQA}$.

For human evaluation, we recruit 45 participants(30 males, 15 females, aged 20-40) from diverse fields (engineering, chemistry, machine learning, and economics). Each participant completes a questionnaire containing 3 randomly selected benchmark samples. For test samples outside an annotator's expertise, we suggest that annotators consult authoritative sources for unfamiliar terminology or background context to better judge whether the explanations are factually grounded.
To ensure consistency across participants, we provided clear instructions and illustrative examples prior to the evaluation. For each task, participants were presented with multiple explanations generated by different methods and asked to make a decision based on each one independently. Immediately afterward, participants selected the explanation they perceived as least cognitively demanding, most logically sound, and most actionable, with the order of explanations randomized for each participant to ensure fairness and minimize bias. This procedure enabled us to analyze trends with statistical confidence.

We define cognitive load as the perceived mental effort required to complete a task. Importantly, actionability and ACC capture distinct concepts. Actionability reflects how easily a third party can select an answer based on an explanation, emphasizing its clarity and usability. By contrast, ACC denotes the accuracy with which a third party can predict the underlying decision based on the explanation. Notably, an explanation may be highly clear (high actionability) yet still lead to an incorrect prediction (low ACC). The score for each metric is computed as the proportion of participants selecting a given method as the best. The example of a sample is listed below. We make sure that all the information about the participants is invisible, and there is no potential risk to their privacy.

\begin{tcolorbox}[title=An example that the evaluation of the effectiveness of our method on $\mathtt{MathQA}$. The explanations generated by $\mathtt{DPO}$; $\mathtt{ReFT}$ and our method are named as A1; A2; A3 and A4 in a random order.]
    \textbf{Description} = Optimal MathQA Problem-Solving Strategies: Choose the Best Explanation: Below is a question along with three explanations (A1, A2, A3). Please review the question and the three explanations carefully, then answer the following questions accordingly\\
    
    \textbf{Q} = A shopkeeper sold an article offering a discount of 5\% and earned a profit of 31.1\%. what would have been the percentage of profit earned if no discount had been offered? A: 38, B: 27.675, C: 30, D: Data Inadequate, E: None of These.
\\
    
    \textbf{A1}: Let's assume that the cost price of the article is 'C'. When the shopkeeper offered a discount of 5\%, the selling price of the article became 95\% of the original selling price. Let's call the original selling price 'S'. Therefore, the new selling price is 0.95S. We know that the shopkeeper earned a profit of 31.1\%. Therefore, we can write the equation: Profit percentage = (Profit / Cost price) x 100. 31.1 = [(0.95S - C) / C] x 100. Now, let's assume that the shopkeeper did not offer any discount. In that case, the selling price would have been the original selling price 'S'. Let's call the new profit percentage 'P'. Therefore, we can write the equation: P = [(S - C) / C] x 100. We need to find the value of 'P' in terms of 'C' and 'S'. From the first equation, we can solve for 'C': C = (0.95S - 0.311S) / 0.311. C = 2.978S. Substituting this value of 'C' in the second equation, we get: P = [(S - 2.978S) / 2.978S] x 100 P = 33.1\%.\\

    After reading Explanation A1, which answer do you think it is leading to?  A: 38; B: 27.675; C: 30; D: Data Inadequate; E: None of These. F: Not Sure\\

    \textbf{A2}: xxx\\
    
    After reading Explanation A2, which answer do you think it is leading to?  A: 38; B: 27.675; C: 30; D: Data Inadequate; E: None of These. F: Not Sure\\

     \textbf{A3}: xxx\\
    
    After reading Explanation A3, which answer do you think it is leading to?  A: 38; B: 27.675; C: 30; D: Data Inadequate; E: None of These. F: Not Sure\\

     \textbf{A4}: xxx\\
    
    After reading Explanation A4, which answer do you think it is leading to?  A: 38; B: 27.675; C: 30; D: Data Inadequate; E: None of These. F:Not Sure\\
    
   Which explanation do you think is the most logically sound? A: A1; B: A2; C: A3; D: A4\\

   Which explanation induce the least cognitive load? A: A1; B: A2; C: A3; D: A4\\

   Which explanation is the most actionable? (Knowing which answer to select after reading it.) A: A1; B: A2; C: A3; D: A4\\

\end{tcolorbox}

% \textbf{AUC}. We calculate the AUC by computing the probability that the score of positive samples exceeds the score of negative samples. To classify the ground-truth positive and negative samples, we use o1-mini to identify the correctness of the explanation generated by the \elm. Subsequently, we employ the \rlm~to assign a score to each explanation generated by the \elm. 

\section{Baselines}\label{appx: baselines}

\textbf{$\mathtt{SFT}$} is a widely used approach to fine-tune LLMs for specific downstream tasks by training them on labeled datasets. In this process, a pre-trained LLM is exposed to task-specific data containing input-output pairs. The goal is to minimize the difference between the model’s predictions and the ground truth, typically using a supervised learning objective like cross-entropy loss. 
We use a training dataset consisting of preferred examples generated by o1-mini. These examples are effective explanations. We use $\mathtt{SFT}$ as the initialization stage to enable the \elm~to adhere to the target format and to set a foundation for subsequent refinement toward the target explanatory behavior. We use BF16 and DeepSpeed ZeRO-2 to accelerate training. The batchsize is 12 and the optimizer is Adam. The training epochs of $\mathtt{SFT}$ is 3 and the learning rate is initialized as $2e\shortn 5$ recommended by the TRL library and then decays linearly to zero.

\textbf{$\mathtt{PPO}$} is a reinforcement learning algorithm commonly used to fine-tune LLMs, enhancing their alignment with specific goals or human preferences. $\mathtt{PPO}$ fine-tunes a pre-trained model by optimizing its policy based on feedback signals. These signals typically come from a reward model, which is trained on labeled data that reflects desirable outputs. $\mathtt{PPO}$ adjusts the model’s parameters to maximize expected reward while constraining updates within a predefined range, avoiding large deviations that could destabilize training. We initialize the model with the same $\mathtt{SFT}$ procedure, leveraging the same o1-mini–distilled preferred examples. To generate more effective explanations, we further fine-tune this $\mathtt{SFT}$ model with a reinforcement learning algorithm, $\mathtt{PPO}$, which is supposed to reinforce explanations that better support inferring the policy's decision by assigning higher rewards.
To achieve this, we train a reward model on a paired preference dataset (queries with corresponding positive/negative explanations). We then fine-tune the $\mathtt{SFT}$ model with $\mathtt{PPO}$ using this reward model. We use BF16 and LoRA with hyperparameters $r_{lora}=16$ and $\alpha_{lora}=16$, while DeepSpeed ZeRO-3 is employed to accelerate training. The batchsize is 12 and the optimizer is Adam. The training epochs of the reward model and $\mathtt{PPO}$ are both 10 and the learning rate is initialized as $3e\shortn 6$ recommended by the TRL library and then decays linearly to zero. The main difference between $\mathtt{PPO}$ and our method lies in the reward modeling. Our method trains a probabilistic reward model using flow matching, while $\mathtt{PPO}$ uses a standard deterministic reward model trained with the Bradley-Terry model.

\textbf{$\mathtt{DPO}$} is a technique for aligning LLMs with human preferences by directly optimizing their outputs using labeled preference data. 
Unlike traditional reinforcement learning from human feedback, which relies on a reward model to evaluate responses, $\mathtt{DPO}$ simplifies the alignment process by directly using preference comparisons to guide optimization. In $\mathtt{DPO}$, the labeled data consists of paired responses where one option is preferred over the other. The model learns to produce outputs that align with these preferences by optimizing a contrastive objective. 
For $\mathtt{DPO}$, we initialize the model with the same $\mathtt{SFT}$ procedure, leveraging the same preferred examples distilled from o1-mini. To generate more effective explanations, we then fine-tune this $\mathtt{SFT}$ model using preference-based learning via $\mathtt{DPO}$. Unlike $\mathtt{PPO}$, $\mathtt{DPO}$ does not require a separate reward model. Instead, we use paired preference data (each query paired with a preferred and a dispreferred explanation), freeze the $\mathtt{SFT}$ checkpoint as the reference model, and fine-tune the \elm~using the $\mathtt{DPO}$ objective. After training converges, we generate explanations using the same prompts as the other baselines. We use BF16 and LoRA with hyperparameters $r_{lora}=16$ and $\alpha_{lora}=16$, while DeepSpeed ZeRO-2 is employed to accelerate training. The batchsize is 12 and the optimizer is Adam. The training epochs of $\mathtt{DPO}$ is 10 and the learning rate is initialized as $5e\shortn 6$ recommended by the TRL library and then decays linearly to zero.

\textbf{$\mathtt{KTO}$} is an advanced approach for aligning LLMs with human preferences or specific task objectives. It draws inspiration from prospect theory, a behavioral economics framework that models how humans evaluate potential gains and losses under uncertainty. In this context, $\mathtt{KTO}$ optimizes the alignment process by weighting outputs based on their perceived utility, rather than treating all errors equally. The core idea of $\mathtt{KTO}$ is to model alignment as an optimization problem where the goal is to maximize expected utility under a prospect-theoretic framework. 
For $\mathtt{KTO}$, we initialize the model using the same $\mathtt{SFT}$ procedure, leveraging the same preferred examples distilled from o1-mini. To further improve explanation quality, we fine-tune the $\mathtt{SFT}$ model using preference-based learning via $\mathtt{KTO}$. Unlike $\mathtt{PPO}$, $\mathtt{KTO}$ does not require a separate reward model. Instead, we construct a dataset of (query, explanation, label) triples, where the label indicates whether the explanation is preferred or dispreferred. We then optimize the \elm~using the $\mathtt{KTO}$ loss, with the $\mathtt{SFT}$ checkpoint kept as a frozen reference for regularization. After training, the $\mathtt{KTO}$-tuned model is prompted with the same context as the other baselines to generate explanations.
We use BF16 and LoRA with hyperparameters $r_{lora}=16$ and $\alpha_{lora}=16$, while DeepSpeed ZeRO-2 is employed to accelerate training. The batchsize is 12 and the optimizer is Adam. The training epochs of $\mathtt{KTO}$ is 10 and the learning rate is initialized as $5e\shortn 7$ recommended by the TRL library and then decays linearly to zero.

\textbf{$\mathtt{Skywork}$}. Skywork-Reward-Llama-3.1-8B-v0.2 is a reward model developed by Skywork AI as part of their efforts to improve alignment and instruction-following abilities in large language models. Based on the LLaMA-3.1 8B architecture, this version—v0.2—is specifically trained to evaluate and rank responses generated by language models in a preference-based learning setting. The model is fine-tuned using a dataset of paired responses labeled by humans or heuristic methods, where each pair contains a preferred and a less preferred answer. This enables the model to predict a scalar reward score for any given response, facilitating its use in fine-tuning chat models for improved helpfulness, relevance, and alignment with human expectations. It ranks among the top two 8B models on the RewardBench Leaderboard, and all top five 8B models adopt it as their backbone and utilize its preference dataset for training. 
For $\mathtt{Skywork}$, we adopt Skywork-Reward-Llama-3.1-8B-v0.2 as the base model. This checkpoint has already been optimized for reward modeling. We further fine-tune it using $\mathtt{SFT}$ on the same preferred explanation dataset to adapt it to the explanation generation task. Then, we apply $\mathtt{PPO}$ using this trained reward model to further align the model's outputs with human preferences. We using BF16 and LoRA with hyperparameters $r_{lora}=16$ and $\alpha_{lora}=16$. DeepSpeed ZeRO-3 is employed to accelerate training. The batchsize is 12 and the optimizer is Adam. The training epoch of $\mathtt{SFT}$ is 3. The training epoch of the reward model and $\mathtt{PPO}$ is both 10, and the learning rate is initialized as $3e\shortn 6$ recommended by the TRL library and then decays linearly to zero.

\textbf{$\mathtt{Deepseek}$}. DeepSeek-R1-Distill-Llama-8B is a high-performance, open-source language model developed by DeepSeek. This model is a distilled version of DeepSeek-R1, built upon the Llama-3.1-8B-Instruct architecture. 
By distilling knowledge from the larger DeepSeek-R1 model, it provides state-of-the-art performance with reduced computational requirements, making it more accessible for various applications. 
For $\mathtt{DeepSeek}$, we use DeepSeek-R1-Distill-Llama-8B as the base model and apply $\mathtt{SFT}$ using the same preferred explanation dataset to adapt it to the explanation generation task.
We use BF16 and LoRA with hyperparameters $r_{lora}=16$ and $\alpha_{lora}=16$. DeepSpeed ZeRO-3 is employed to accelerate training. The batchsize is 12 and the optimizer is Adam. The training epoch of the $\mathtt{SFT}$ model is 3 and the learning rate is initialized as $2e\shortn 5$ recommended by the TRL library and then decays linearly to zero.

\textbf{$\mathtt{CoT}$+$\mathtt{SFT}$} is a fine-tuning method that enhances the reasoning capabilities of LLMs by combining $\mathtt{SFT}$ with the structured reasoning paradigm. CoT uses explicit programmatic representations, such as pseudo-code or structured logic, to model complex problem-solving tasks. In this approach, $\mathtt{SFT}$ is performed using datasets annotated with both input-output pairs and detailed programmatic reasoning traces. These traces serve as templates for step-by-step reasoning and enable the model to break down complex problems, such as mathematical reasoning or logical inference, into manageable sub-tasks. The explicit program-like structure helps the model perform multi-step computations and enhances interpretability, making it especially useful for domains requiring precision and transparency. 
For $\mathtt{CoT}$+$\mathtt{SFT}$, we construct a $\mathtt{CoT}$-augmented version of our preferred explanation dataset via o1-mini by replacing each explanation with a $\mathtt{CoT}$ variant, containing more explicit step-by-step reasoning. We then apply $\mathtt{SFT}$ using this modified dataset to the same base model as our method. We adopt the open source library from
\citep{trung2024reft}, where we use BF16, DeepSpeed ZeRO-2, a fixed learning rate of 1e-5, Adam optimizer and a batchsize of 12.

\textbf{$\mathtt{ReFT}$} is a training approach designed to enhance the reasoning capabilities of LLMs by combining supervised fine-tuning ($\mathtt{SFT}$) with reinforcement learning. In $\mathtt{ReFT}$, the initial training begins with $\mathtt{SFT}$, where the model is fine-tuned using datasets annotated with reasoning traces, such as step-by-step explanations or logical chains of thought. Once the model achieves a baseline performance, reinforcement learning is applied to further refine its reasoning capabilities. 
For $\mathtt{ReFT}$, we use the same $\mathtt{COT}$-augmented dataset as in $\mathtt{CoT}$+$\mathtt{SFT}$ and the $\mathtt{ReFT}$ algorithm to the same base model as our method. We adopt the open source library from \citep{trung2024reft}, where we use BF16, DeepSpeed ZeRO-2, a fixed learning rate of 3e-7,  Adam optimizer and a batchsize of 12. 

\textbf{$\mathtt{o3}$-$\mathtt{mini}$} is a smaller, more cost-efficient reasoning model in OpenAI’s “o3” family. It’s optimized for tasks in STEM domains — especially math, science, and coding — with the goal of higher accuracy, lower latency, and lower cost compared to larger models. We directly use it for evaluation without fine-tuning.

$\mathtt{o3}$-$\mathtt{mini}$ is used exclusively for inference without fine-tuning, whereas $\mathtt{CoT}$+$\mathtt{SFT}$ and $\mathtt{ReFT}$ rely on the $\mathtt{CoT}$-enhanced dataset. All other baselines, as well as our method, are trained on the same dataset.

\section{Additional Results}\label{appx:results}

\subsection{Different Choices of $\sigma$ in $q_0$}

We evaluate the impact of varying $\sigma$ in $q_0$ on final performance across different datasets. As shown in Table \ref{tab: different_sigma}, the results remain stable across a wide range of values ($\sigma^2 = 0.5, 1, 5$). Specifically, on SMAC, the performance slightly decreases from 0.772 at $\sigma^2=0.5$ to 0.761 at $\sigma^2=5$, while MMLU exhibits a small improvement from 0.759 to 0.781 as $\sigma^2$ increases. For MathQA, performance is highest at $\sigma^2=0.5$ (0.809) and decreases mildly as $\sigma^2$ grows. These variations are minimal, underscoring that our method is not sensitive to the choice of $\sigma$ and thereby confirming the robustness predicted by our theoretical analysis.
\begin{table} [h]
    \caption{Performance of our method with different $\sigma$ in $q_0$.}
    \centering
    \scalebox{0.95}{
    \vspace{-0.8em}
    
    \begin{tabular}{crcrcrcrcrcrcr}
        \toprule
        
        \multicolumn{2}{l}{Datasets} &
        \multicolumn{2}{c}{$\sigma^2=0.5$}&
        \multicolumn{2}{c}{$\sigma^2=1$}&
        \multicolumn{2}{c}{$\sigma^2=5$}\\

         % \midrule

         \cmidrule(lr){1-2}
        \cmidrule(lr){3-4}
        \cmidrule(lr){5-6}
        \cmidrule(lr){7-8}

           % \cmidrule(lr){19-20}
        % \cmidrule(lr){21-22}

        % \cmidrule(lr){5-14}

        % \cmidrule(lr){1-2}
        % \cmidrule(lr){3-4}
        % \cmidrule(lr){5-6}
        % \cmidrule(lr){7-8}
        % \cmidrule(lr){9-10}
        % \cmidrule(lr){11-12}
        % \cmidrule(lr){13-14}

        % & \multicolumn{2}{l}{100\%} & \multicolumn{2}{l}{100\%} & \multicolumn{2}{l}{100\%} & \multicolumn{2}{l}{100\%} \\
        % \multicolumn{2}{c}{} & \multicolumn{2}{l}{}

        \multicolumn{2}{c}{${\mathtt{SMAC}}$} & 
        \multicolumn{2}{c}{0.772} & 
        \multicolumn{2}{c}{0.764} & 
        \multicolumn{2}{c}{0.761}   \\
        
       \cmidrule(lr){1-2}
        \cmidrule(lr){3-4}
        \cmidrule(lr){5-6}
        \cmidrule(lr){7-8}

        % \cmidrule(lr){21-22}

        \multicolumn{2}{c}{${\mathtt{MMLU}}$
        } & 
        \multicolumn{2}{c}{0.759}  & 
        \multicolumn{2}{c}{0.772} & 
        \multicolumn{2}{c}{0.781}  \\

        \cmidrule(lr){1-2}
        \cmidrule(lr){3-4}
        \cmidrule(lr){5-6}
        \cmidrule(lr){7-8}

           % \cmidrule(lr){19-20}
        % \cmidrule(lr){21-22}

        \multicolumn{2}{c}{${\mathtt{MathQA}}$} & 
        \multicolumn{2}{c}{0.809}  & 
        \multicolumn{2}{c}{0.804} & 
        \multicolumn{2}{c}{0.783}  \\

        \toprule
    \end{tabular}}
    % \begin{tablenotes}
    %     \item[1] xxx.
    % \end{tablenotes}
    \label{tab: different_sigma}
    \vspace{-0.5em}
\end{table}
%sigma@2=1，0.5，2

\subsection{Without SFT}
To evaluate the influence of supervised fine-tuning on our method, we train our method and baselines without conducting $\mathtt{SFT}$. Results are shown in Tab.~\ref{tab: wosft}. Our method does require $\mathtt{SFT}$ to achieve strong performance, possibly because $\mathtt{SFT}$ data is crucial for obtaining enough positive samples to train the rectified flow model. Removing SFT, the performance of our method and baselines drops particularly sharply on RL tasks like SMAC. RL tasks typically involve sequential decision-making and even interactions among multiple agents, making the reasoning of correct actions particularly challenging. $\mathtt{SFT}$ offers valuable examples about how to reason in such complex scenarios. Notably, our method without $\mathtt{SFT}$ significantly outperforms baselines without $\mathtt{SFT}$ on $\mathtt{SMAC}$, demonstrating its effectiveness in cases characterized by higher reasoning demands.

\begin{table} [h]
    \caption{Performance of our method and baselines without supervised fine-tuning.}
    \centering
    \scalebox{0.95}{
    \vspace{-0.8em}
    
    \begin{tabular}{crcrcrcrcrcrcr}
        \toprule
        
        \multicolumn{2}{l}{Datasets} &
        \multicolumn{2}{c}{${\mathtt{Ours}}$}&
        \multicolumn{2}{c}{${\mathtt{PPO}}$}&
        \multicolumn{2}{c}{${\mathtt{DPO}}$}&
        \multicolumn{2}{c}{${\mathtt{KTO}}$}&
        \multicolumn{2}{c}{${\mathtt{ReFT}}$}&
        \multicolumn{2}{c}{${\mathtt{Skywork}}$}\\

         % \midrule

         \cmidrule(lr){1-2}
        \cmidrule(lr){3-4}
        \cmidrule(lr){5-6}
        \cmidrule(lr){7-8}
           \cmidrule(lr){9-10}
        \cmidrule(lr){11-12}
        \cmidrule(lr){13-14}

           % \cmidrule(lr){19-20}
        % \cmidrule(lr){21-22}

        % \cmidrule(lr){5-14}

        % \cmidrule(lr){1-2}
        % \cmidrule(lr){3-4}
        % \cmidrule(lr){5-6}
        % \cmidrule(lr){7-8}
        % \cmidrule(lr){9-10}
        % \cmidrule(lr){11-12}
        % \cmidrule(lr){13-14}

        % & \multicolumn{2}{l}{100\%} & \multicolumn{2}{l}{100\%} & \multicolumn{2}{l}{100\%} & \multicolumn{2}{l}{100\%} \\
        % \multicolumn{2}{c}{} & \multicolumn{2}{l}{}

        \multicolumn{2}{c}{${\mathtt{SMAC}}$} & 
        \multicolumn{2}{c}{0.457} & 
        \multicolumn{2}{c}{0.302} & 
        \multicolumn{2}{c}{0.324} & 
        \multicolumn{2}{c}{0.381} & 
        \multicolumn{2}{c}{0.405} & 
        \multicolumn{2}{c}{0.357}  \\
        
       \cmidrule(lr){1-2}
        \cmidrule(lr){3-4}
        \cmidrule(lr){5-6}
        \cmidrule(lr){7-8}
           \cmidrule(lr){9-10}
        \cmidrule(lr){11-12}
        \cmidrule(lr){13-14}

        % \cmidrule(lr){21-22}

        \multicolumn{2}{c}{${\mathtt{MMLU}}$
        } & 
        \multicolumn{2}{c}{0.605}  & 
        \multicolumn{2}{c}{0.586} & 
        \multicolumn{2}{c}{0.590} & 
        \multicolumn{2}{c}{0.621} & 
        \multicolumn{2}{c}{0.602} & 
        \multicolumn{2}{c}{0.608} \\

        \cmidrule(lr){1-2}
        \cmidrule(lr){3-4}
        \cmidrule(lr){5-6}
        \cmidrule(lr){7-8}
           \cmidrule(lr){9-10}
        \cmidrule(lr){11-12}
        \cmidrule(lr){13-14}

           % \cmidrule(lr){19-20}
        % \cmidrule(lr){21-22}

        \multicolumn{2}{c}{${\mathtt{MathQA}}$} & 
        \multicolumn{2}{c}{0.634}  & 
        \multicolumn{2}{c}{0.618} & 
        \multicolumn{2}{c}{0.630} & 
        \multicolumn{2}{c}{0.638} & 
        \multicolumn{2}{c}{0.643} & 
        \multicolumn{2}{c}{0.629}  \\

        \toprule
    \end{tabular}}
    % \begin{tablenotes}
    %     \item[1] xxx.
    % \end{tablenotes}
    \label{tab: wosft}
    \vspace{-0.5em}
\end{table}

\subsection{Under Noisy Reward}
To evaluate the robustness of our method under noisy reward signals, we conduct an additional experiment where the reward (0/1) of the final token is randomly flipped with probability 0.03. As shown in Tab.~\ref{tab: noise_reward}, under the noisy reward condition, the performance of our method decreases only slightly from 77.2\% to 74.3\% on unseen tasks in the $\mathtt{MMLU}$ benchmark. Notably, even with these noisy rewards, our method still outperforms $\mathtt{PPO}$ (72.3\%), $\mathtt{Skywork}$ (73.7\%), and $\mathtt{DeepSeek}$ (72.1\%) trained without noise. This demonstrates the robustness of our approach under imperfect reward feedback.

\begin{table} [h]
    \caption{Performance of our method under noisy reward.}
    \centering
    \scalebox{0.95}{
    
    \begin{tabular}{crcrcrcr}
        \toprule        
        \multicolumn{2}{l}{} &
      \multicolumn{2}{c}{SMAC}&\multicolumn{2}{c}{MMLU}&\multicolumn{2}{c}{MathQA}\\

         \cmidrule(lr){1-2}
        \cmidrule(lr){3-4}
        \cmidrule(lr){5-6}
        \cmidrule(lr){7-8}

         \multicolumn{2}{l}{$\mathtt{Ours}$ $\mathtt{(Noise)}$} & 
        \multicolumn{2}{c}{$\backslash$} & 
        \multicolumn{2}{c}{0.743}& 
        \multicolumn{2}{c}{$\backslash$} 
        
         \\

        \toprule
    \end{tabular}}
   
    \label{tab: noise_reward}
  
\end{table}

\subsection{Visualized Examples}

\textbf{$\mathtt{SMAC}$}. To showcase the effectiveness of our method, we visualize an example on $\mathtt{SMAC}$ that a negative explanation classified by \rlm~is corrected by the rectified flow model $\varphi$. As shown above, this explanation infers the answer is related to `Attack Enemy 0', however, the \rlm~fails to predict the correct action. Instead, the rectified flow model $\varphi$ can predict the action correctly based on the explanation.

Besides, we visualize another example that a negative explanation generated by \elm~is optimized to a positive explanation after training for two rounds. As shown below, the explanation from the ${\mathtt{SFT}}$ model opts for the movement of EAST. The explanation of our method can infer the correct movement of NORTH with solid evidence.

\textbf{$\mathtt{MathQA}$}. To showcase the effectiveness of our method, we visualize an example on $\mathtt{MathQA}$ that a negative explanation classified by \rlm~is corrected by the rectified flow model $\varphi$. As shown below, this explanation infers the answer is related to option d, however, the \rlm~fails to predict the correct answer. Instead, the rectified flow model $\varphi$ can predict the answer correctly based on the explanation.

Besides, we visualize another example on $\mathtt{MathQA}$ that a negative explanation generated by \elm~is optimized to a positive explanation after training for two rounds. As shown below, the explanation from the ${\mathtt{SFT}}$ model is confusing and misleading. The explanation of our method can infer the correct answer c with solid evidence.

\begin{tcolorbox}[title=An example that a negative explanation classified
by the \rlm~is corrected by the rectified flow model $\varphi$ of on $\mathtt{SMAC}$]

    \textbf{Q} = Here are 5 allies and 6 enemies in SMAC. The goal of each ally is to attack all the enemies together with other allies. At every step, each ally can choose one action to take from the Action set. The trajectory of each ally consists of a series of State, Action, Value pairs. Here are 4 consecutive State, Action, Value pairs where the last state-action pair misses the Action: xxx. Please help me concisely summarize and analyze this part of the trajectory for the missing Action prediction.\\
    
    \textbf{Action set $\mathcal{A}$} = [`DEAD', `STOP', `NORTH', `SOUTH', `EAST', `WEST', \textcolor{outputcolor} {`Attack Enemy 0' $\checkmark$}, `Attack Enemy 1', `Attack Enemy 2', `Attack Enemy 3', `Attack Enemy 4', `Attack Enemy 5']\\
    % In the provided trajectory, each ally demonstrates a cautious yet proactive approach to engaging enemies. 
    
    \textbf{{Explanation:}} ...In the first state, Ally opts to move EAST, likely positioning itself xxx. Ally’s next action, “Attack Enemy 3”, shows a shift to offensive behavior, ... \textcolor{outputcolor}{Ally’s following action, “Attack Enemy 0”, further prioritizes offense, which might reflect a high-risk engagement. Ally’s last action solidifies this pattern, further engaging Enemy 0, likely due to the imminent engagement's hazards. Throughout these actions, ... The allies prioritize attacking specific foes based on their availability and proximity.$\checkmark$}  \\
    
    \textbf{{Distribution $p$ from the \rlm:}} [0.0074, 0.0031, 0.0260, 0.0062, 0.0055, 0.0096, 0.1069, 0.2148, \textbf{0.2812}, 0.1089, 0.1260, 0.1045] $\rightarrow$ `Attack Enemy 2’ {\color{red}$\boldsymbol{\times}$}\\\
    
    \textbf{{Distribution $\hat{p}$ from the rectified flow model:}} 
[0.0084, 0.0040, 0.0084, 0.0061, 0.0064, 0.0099, \textbf{0.3254}, 0.1110, 0.1137, 0.1771, 0.1065, 0.1218] $\rightarrow$ `Attack Enemy 0’ \textcolor{outputcolor} {$\checkmark$}\\

\end{tcolorbox}

\begin{tcolorbox}[title=An example on $\mathtt{SMAC}$ that a negative explanation generated by the \rlm~is optimized to a positive explanation after training for two rounds]

\textbf{Q} = Here are 5 allies and 6 enemies in $\mathtt{SMAC}$. The goal of each ally is to attack all the enemies together with other allies. At every step, each ally can choose one action to take from the Action set. The trajectory of each ally consists of a series of State, Action, Value pairs. Here are 4 consecutive State, Action, Value pairs where the last state-action pair misses the Action: xxx. Please help me concisely summarize and analyze this part of the trajectory for the missing Action prediction.\\

    \textbf{Actual Action} = NORTH\\
    % In the provided trajectory, each ally demonstrates a cautious yet proactive approach to engaging enemies. 
    
    \textbf{{Explanation:}} In the first state, ally 0,1,2, and 3 all move south, ... The following action, STOP, ... Subsequently, ally 0 and 2 move WEST, ... Throughout these steps, allies prioritize movement over direct engagement, likely because no enemies are in view. Given the current state, where allies are arranged with their backs touching and no Enemy is in sight, the next logical action is EAST. This directional movement likely aims to realign the formation back to its original or a more advantageous state, ensuring that allies are positioned optimally for future engagements. {\color{red}This EAST movement contrasts with WEST, which may have been a short-term repositioning to balance the formation }{\color{red}(EAST, $\boldsymbol{\times}$)}, and NORTH and SOUTH, which were likely attempts to regroup or reposition relative to other allies. \\
    
    \textbf{{Improved Explanation (Ours):}}   In the first state, the ally chooses to move SOUTH, ... In the second state, the ally STOPs, ... Moving WEST in the third state likely aims to cover more ground or flank the enemies, ... In the latest state, with no visible enemies and all allies maintaining full health and no weapon cooldowns, the ally opts to move NORTH. This directional movement decision likely stems from a strategic need to reposition for a better vantage point or to avoid potential threats, \textcolor{outputcolor} {given the NORTHward movement may offer a more advantageous stance compared to other directions.$\checkmark$} Choosing NORTH over EAST, for instance, may be due to factors such as better cover or better line-of-sight, enhancing the ally's readiness to engage when enemies become visible.  \\

\end{tcolorbox}

\begin{tcolorbox}[title=An example that a negative explanation classified
by the \rlm~is corrected by the rectified flow model $\varphi$ of on $\mathtt{MathQA}$]
    \textbf{Q} =  the cash realised on selling a 14\% stock is rs. 101.25, brokerage being 1/4\% is.\\
    
    \textbf{Decision set $\mathcal{A}$} = [a: 123, b: 106, c: 100, \textcolor{outputcolor}{d: 101 $\checkmark$}, e: 240]\\
    % In the provided trajectory, each ally demonstrates a cautious yet proactive approach to engaging enemies. 
    
    \textbf{{Explanation:}} cash realised = rs. \textcolor{outputcolor}{( 101.25 - 0.25 )$\checkmark$} = rs \\
    
    \textbf{{Distribution $p$ from the \rlm:}} [0.1035, 0.0481, \textbf{0.6289}, 0.2061, 0.0122] $\rightarrow$ `c’ {\color{red}$\boldsymbol{\times}$}\\\
    
    \textbf{{Distribution $\hat{p}$ from the rectified flow model::}} 
[0.2363, 0.0477, 0.0114, \textbf{0.6683}, 0.0567] $\rightarrow$ `d’ \textcolor{outputcolor} {$\checkmark$}\\

\end{tcolorbox}

\begin{tcolorbox}[title=An example on $\mathtt{MathQA}$ that a negative explanation generated by the \rlm~is optimized to a positive explanation after training for two rounds]

    \textbf{Q} = A pet store regularly sells pet food at a discount of 10 percent to 30 percent from the manufacturer’s suggested retail price. if during a sale, the store discounts an additional 20 percent from the discount price, what would be the lowest possible price of a container of pet food that had a manufacturer’s suggested retail price of \$ 40.00?\\

    \textbf{Decision set $\mathcal{A}$} = [a: \$ 10.00, b: \$ 11.20, \textcolor{outputcolor}{c: \$ 22.40 $\checkmark$}, d: \$ 16.00, e: \$ 18.00.]\\
    % In the provided trajectory, each ally demonstrates a cautious yet proactive approach to engaging enemies. 
    
    \textbf{{Explanation:}} retail price = manufacturer suggested retail price = \$ 40 first discount price = retail price {\color{red}-10 \% to -30 \% = presence of retail price retrievable.$\boldsymbol{\times}$}   \\
    
    \textbf{{Improved Explanation (Ours):}} retail price =40 first maximum usually discounted price =40 - 30 \% of 40 =40 - 12 =28 price after additional discounting during sale =\textcolor{outputcolor}{28 - 20 \% of 28 =28 - 5.60 =22.40 $\checkmark$} \\

\end{tcolorbox}

\section{Limitation}

While the flow model is theoretically grounded and performs well in capturing reward distributions aligned with human preferences, it may still face generalization challenges in edge cases—such as ambiguous preference data, rare or unseen scenarios, and sensitive topics—similar to other LLM-based reward models. 

Besides, our method assumes that decisions are context-dependent. For random actions, which lack a consistent rationale, the model cannot generate faithful explanations.

Moreover, learning from human feedback and ethical reasoning are two examples of such tasks that involve multiple plausible choices. Human feedback and ethical judgments may vary with different individuals, and the viewpoints of a population together form a complex distribution. Our method is suitable for such learning tasks. By contrast, our method may be less effective in domains such as abstract knowledge, where producing logical or factual rules is required. In such cases, constructing meaningful rewards from \rlm~s becomes more challenging. 

Although our design choice foregoes direct analysis of internal mechanisms, we think that this pragmatic trade-off is warranted when developing methods intended to generalize across model classes with fundamentally different or opaque internals. It affords two advantages that are central to our goals: (i) it enables a unified treatment of diverse architectures—including RL policies and LLM agents—whose internal structures may be opaque, inaccessible, or highly heterogeneous; and (ii) it guarantees behavioral (functional) faithfulness, which is often a primary concern in safety-critical or deployment settings.

\section{Impact Statements}
This paper presents work whose goal is to advance the field of machine learning by developing a model-agnostic explanation generator for intelligent agents, enhancing transparency and interpretability in agent decision prediction. The ability to generate effective and interpretable explanations has the potential to foster trust in AI systems, improving effectiveness in high-stakes applications such as healthcare, finance, and autonomous systems. Overall, we believe our work contributes positively to the broader AI ecosystem by promoting more explainable and trustworthy AI.

\end{document}